\pdfoutput=1
\documentclass{ieeetj}
\usepackage{cite}
\usepackage{amsmath,amssymb,amsfonts}
\usepackage{algorithm}
\usepackage{graphicx,color}
\usepackage{textcomp}
\usepackage{xcolor}
\usepackage[colorlinks = true,
            linkcolor = blue,
            urlcolor  = blue,
            citecolor = blue,
            anchorcolor = blue]{hyperref}
\usepackage{algpseudocode}
\usepackage[beginLComment=/*~,endLComment=~*/]{algpseudocodex}
\usepackage{subcaption}
\usepackage{etoolbox}

\newcommand\crule[3][black]{\textcolor{#1}{\rule{#2}{#3}}}
\definecolor{geomblue}{HTML}{a4c2f4}
\definecolor{visiongreen}{HTML}{B6D7A8}
\definecolor{triangorange}{HTML}{f9cb9c}
\definecolor{fusionred}{HTML}{ea9999}
\definecolor{plannerpurple}{HTML}{b4a7d6}
\definecolor{jetblue}{HTML}{030bff}
\definecolor{jetred}{HTML}{ff0303}
\definecolor{edgeRed}{HTML}{D62728}
\definecolor{freeGreen}{HTML}{10ff03}
\definecolor{frontierBlue}{HTML}{030bff}
\definecolor{scoreJet}{HTML}{ff0303}
\definecolor{cyanGoal}{HTML}{03d9ff}
\definecolor{particleWhite}{HTML}{E6E6E6}
\definecolor{pathGreen}{HTML}{10ff03}

\usepackage{colortbl}
\usepackage{booktabs}
\usepackage{array}
\usepackage{multirow}
\usepackage{siunitx}

\usepackage{pifont}
\newcommand{\xmark}{\ding{55}}%

\DeclareMathOperator*{\argmin}{argmin}

\def\BibTeX{{\rm B\kern-.05em{\sc i\kern-.025em b}\kern-.08em
    T\kern-.1667em\lower.7ex\hbox{E}\kern-.125emX}}
\AtBeginDocument{\definecolor{tmlcncolor}{cmyk}{0.93,0.59,0.15,0.02}\definecolor{NavyBlue}{RGB}{0,86,125}}

\def\OJlogo{\vspace{-4pt}{\rffont IEEE Transactions on Field Robotics}}
\def\seclogo{\vspace{10pt}{\rffont IEEE Transactions on Field Robotics}}

\makeatletter
\patchcmd{\@maketitle}{\fi.\space}{\fi\space}{}{\PackageError{main}{Could not remove date-line period}{}}
\let\@revised\@empty
\def\ps@plain{%
   \def\@oddhead{\vbox{\hsize\textwidth\vspace*{-13pt}\vbox to 29pt{\hsize\textwidth\vspace*{11.5pt}\OJlogo}\hfill\par\vspace*{-2.75pt}\hbox to \textwidth{\vrule width\textwidth height.3pt depth0pt}}}%
   \let\@evenhead\@oddhead
   \def\@oddfoot{\hbox to \textwidth{\hfill{\rffont\thepage}}}%
   \let\@evenfoot\@oddfoot}
\def\ps@headings{%
   \def\@oddhead{\vbox{\hsize\textwidth\vspace*{-13.5pt}\hfill\hbox{\vbox to 29pt{\hsize257pt\vspace*{-3pt}\raggedleft\seclogo}}\par\vspace*{-2.25pt}\hbox to \textwidth{\vrule width\textwidth height.3pt depth0pt}}}%
   \def\@evenhead{\vbox{\hsize\textwidth\vbox to 0pt{\hsize\textwidth\vspace*{7.7pt}\rfxfont\raggedright\rightmark:\ \leftmark\hfill\par}\par\vspace*{16pt}\hbox to \textwidth{\vrule width\textwidth height.3pt depth0pt}}}%
   \def\@evenfoot{\hbox to \textwidth{{\rffont\thepage}\hfill}}%
   \def\@oddfoot{\hbox to \textwidth{\hfill{\rffont\thepage}}}}
\makeatother

\def\authorrefmark#1{\ensuremath{^{\textbf{#1}}}}

\begin{document}
\editor{This article has been accepted for publication in IEEE Transactions on Field Robotics. This is the author's accepted version; the final published version may differ.\endgraf\vskip3pt
\textcopyright~2026 IEEE. Personal use of this material is permitted. Permission from IEEE must be obtained for all other uses, in any current or future media, including reprinting/republishing this material for advertising or promotional purposes, creating new collective works, for resale or redistribution to servers or lists, or reuse of any copyrighted component of this work in other works.}
\doiinfo{TFR.2026.3737358}

\markboth{WildOS: Open-Vocabulary Object Search in the Wild}{Shah {et al.}}

\title{WildOS: Open-Vocabulary Object Search in the Wild}

\author{Hardik Shah\authorrefmark{1,2}, Manthan Patel\authorrefmark{2}, Erica Tevere\authorrefmark{1}, Deegan Atha\authorrefmark{3}, Marcel Kaufmann\authorrefmark{1}, Shehryar Khattak\authorrefmark{3}, Marco Hutter\authorrefmark{2}, Jonas Frey\authorrefmark{2,4}, Patrick Spieler\authorrefmark{1}}
    \affil{Jet Propulsion Laboratory (JPL), California Institute of Technology (Caltech), Pasadena, CA 91011, USA}
    \affil{Swiss Federal Institute of Technology (ETH Z\"urich), Robotic Systems Lab, Z\"urich 8092, Switzerland}
    \affil{FieldAI Inc. 3 Morgan, Irvine, CA, USA. Work Performed at NASA JPL}
    \affil{University of California Berkeley, Berkeley, CA 94720, USA}
    \authornote{The research was carried out at the Jet Propulsion Laboratory, California
    Institute of Technology, under a contract with the National Aeronautics and Space Administration (80NM0018D0004). The work was supported in part by the Luxembourg National Research Fund (Ref. 18990533) and the Swiss National Science Foundation (SNSF) as part of the projects No.200021E\_229503 and No.227617. Hardik Shah was supported by the Zeno Karl Schindler Foundation Master's Thesis Grant during this work.}

\begin{abstract}

Autonomous navigation in complex, unstructured outdoor environments requires robots to operate over long ranges without prior maps and with limited depth sensing. In such settings, relying solely on geometric frontiers for exploration is often insufficient, and the ability to reason semantically about \emph{where to go} and \emph{what is safe to traverse} is crucial for robust, efficient exploration. This work presents WildOS, a unified system for long-range, open-vocabulary object search that combines safe geometric exploration with semantic visual reasoning. WildOS builds a sparse navigation graph to maintain spatial memory, while utilizing a foundation-model-based vision module, ExploRFM, to score frontier nodes of the graph. ExploRFM simultaneously predicts traversability, visual frontiers, and object similarity in image space, enabling real-time, onboard semantic navigation tasks. The resulting vision-scored graph enables the robot to explore semantically meaningful directions while ensuring geometric safety. Furthermore, we introduce a multi-view particle-based triangulation method for coarse localization of the open-vocabulary target query, which estimates candidate goal positions beyond the robot’s immediate depth horizon, enabling effective planning toward distant goals. Extensive closed-loop field experiments across diverse off-road and urban terrains demonstrate that WildOS enables robust navigation, outperforming purely geometric and purely vision-based baselines in both efficiency and autonomy. Our results highlight the potential of vision foundation models to drive open-world robotic behaviors that are both semantically informed and geometrically grounded. Project Page: \href{https://leggedrobotics.github.io/wildos/}{leggedrobotics.github.io/wildos}

\end{abstract}


\maketitle
 
\begin{figure*}[t]
    \centering
    \includegraphics[width=\textwidth]{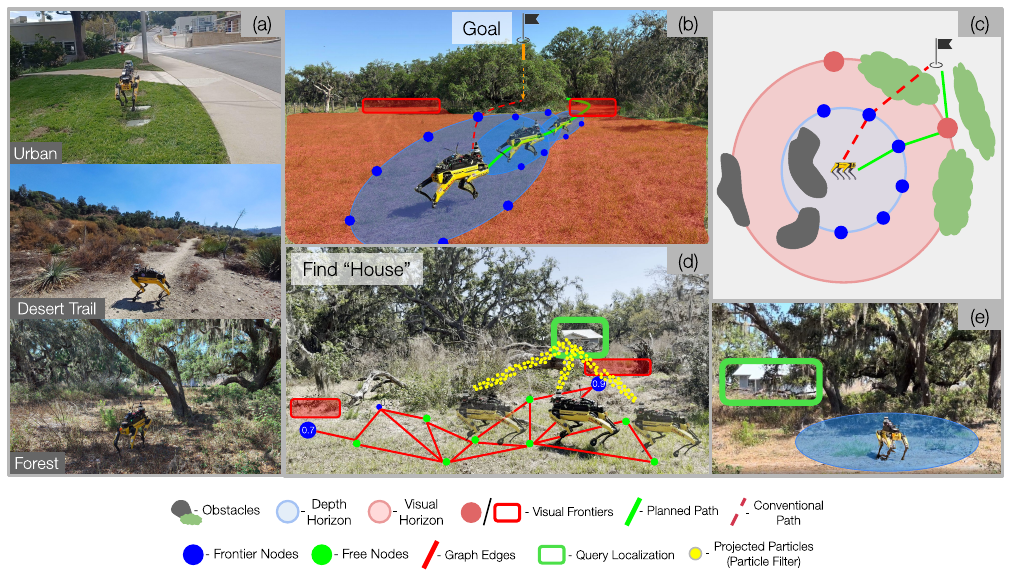}
    \caption{
    \textbf{(a)} WildOS enables autonomous semantic navigation in diverse unstructured outdoor environments. \textbf{(b, c)} Due to the limited range of geometric sensing, robots can only reliably perceive nearby regions within a depth horizon (blue), leading to myopic exploration and \textbf{(e)} difficulty localizing distant targets (e.g., a “house”) beyond sensing range. \textbf{(b, c)} Conventional exploration (dashed path) relies on geometric frontiers (blue dots) at the boundary between known and unknown space, which ignores long-range semantic and traversability cues. WildOS (green path) augments geometric exploration with long-range visual reasoning using a vision foundation model, defining a visual horizon (red) that extends beyond depth sensing and predicts visual traversability, visual frontiers (red dots), and open-vocabulary object similarity in image space. \textbf{(d)} During deployment, a sparse navigation graph is built from geometry and frontier nodes are scored using vision, while a particle-based goal triangulation module (yellow particles) estimates candidate goal locations beyond the depth horizon, enabling safe, efficient planning toward distant semantic goals.
    }
    \label{fig:teaser}
\end{figure*}

\section{INTRODUCTION}
\IEEEPARstart{O}{utdoor} autonomous robots are increasingly being deployed for tasks such as search-and-rescue, remote inspection, and environmental monitoring--settings where prior maps are unavailable and human supervision is limited. In these scenarios, robots must reason not only about \emph{how} to navigate safely, but also \emph{where} to go next in pursuit of semantically meaningful goals. We study this problem in the context of long-range, open-vocabulary object search in large, unstructured environments (Fig.~\ref{fig:teaser}(a)), where a robot must locate and reach a target object described in natural language, such as “find the house,” using only onboard sensing and without global maps.
This problem is challenging for several reasons.  
First, the robot perceives the world only through a \emph{limited local horizon} determined by the sensing range of its depth or LiDAR system. As illustrated in Fig.~\ref{fig:teaser}(b,c), the local metric costmap typically covers only a few meters around the robot. Beyond this range, depth information becomes sparse and noisy, leaving the remainder of the environment unknown. Existing navigation systems~\cite{roadrunner} handle this by heuristically assigning a constant traversal cost to unknown regions and planning straight toward the goal. While effective in simple open terrain, this strategy often leads to myopic, or highly inefficient paths when obstacles--such as fences or undergrowth--block the direct route.

\par
Second, purely geometry-based exploration neglects the \emph{semantic information} readily available in images. Consider Fig.~\ref{fig:teaser}(b,c): beyond the depth horizon, the image provides rich visual cues about open regions or “affordable” gaps in the scene -- what recent work~\cite{schmittle2025lrn} terms \emph{visual frontiers}. These frontiers capture affordances such as openings between obstacles or safe continuations of the path, enabling decisions that resemble human reasoning: preferring to move toward visually navigable regions rather than blindly following the goal direction. However, existing vision-based methods, while effective locally, operate in a memoryless manner and do not retain information about previously explored regions, frequently resulting in oscillations or repeated exploration of the same areas.

\par
Third, the robot must maintain a representation of explored versus unexplored regions over very long ranges. Dense voxel or metric maps are memory-intensive and do not scale well to large outdoor environments. Sparse, graph-based representations, in contrast, provide a memory-efficient structure for maintaining both topological connectivity and exploration history. Yet, previous works have not integrated vision-based reasoning within such a graph framework, limiting their ability to combine geometric safety with long-range semantic awareness.

\par
Finally, a key challenge unique to open-vocabulary \emph{object search} lies in localizing goal objects that may lie far beyond accurate depth sensing. As shown in Fig.~\ref{fig:teaser}(e), the robot can visually detect an object but cannot reliably localize it in 3D space due to the lack of valid range measurements. This prevents existing open-vocabulary mapping methods from providing usable localization for long-range planning.
\newline
Given these challenges, we ask:  
\begin{quote}
    \emph{How can a robot unify visual and geometric reasoning to perform efficient, long-range, open-vocabulary object search in unstructured outdoor environments?}
\end{quote}

\par
To answer this question, we present \textbf{WildOS}-a unified, real-time system that combines the complementary strengths of visual and geometric perception for long-range navigation and object search (Fig.~\ref{fig:teaser}(d)). At its core, WildOS maintains a \emph{navigation graph} that stores explored regions and identifies frontier nodes at the boundary of the known space. Each frontier node is scored using visual information derived from \textbf{ExploRFM}--our foundation-model-based vision module that predicts visual frontiers, traversability, and open-vocabulary object similarity directly in the image space. The resulting \emph{vision-scored graph} allows WildOS to plan through semantically meaningful, visually promising directions while retaining the geometric safety and spatial memory of the graph structure. 
Furthermore, WildOS introduces a multi-view particle-based triangulation approach for coarse 3D localization of open-vocabulary goal objects beyond the robot’s depth horizon. This enables goal-aware planning even when the object lies far outside the depth sensing range.

\par
Through extensive field experiments in unstructured off-road and urban terrains, we demonstrate that WildOS improves efficiency and robustness over both purely geometric and purely vision-based baselines. As summarized in Fig.~\ref{fig:teaser}, our system attempts to naturally exhibit human-like navigation behavior, such as selecting openings between obstacles early, turning back intelligently upon encountering dead-ends, and maintaining consistent progress toward semantically meaningful goals. In summary, we make the following key contributions:

\begin{itemize}
    \item \textbf{WildOS}: A unified, real-time system for long-range, open-vocabulary object search that integrates geometric and visual reasoning through a vision-scored navigation graph.
    \item \textbf{ExploRFM Module}: A foundation-model-based network that jointly predicts traversability, visual frontiers, and object similarity in image space for onboard decision-making.
    \item \textbf{Vision-Scored Graph}: A novel topological mapping approach that scores geometric frontiers with semantic cues, prioritizing exploration toward visually promising regions.
    \item \textbf{Beyond-Horizon Object Localization}: A multi-view particle-based triangulation estimator that produces a coarse position prior for targets situated outside the robot's depth range, enabling goal-aware planning toward distant visual cues.
    \item \textbf{Field Validation \& Dataset}: Extensive closed-loop experiments demonstrating improved performance over SOTA baselines in complex off-road/urban terrain, supported by a new manually annotated dataset for visual frontiers.
\end{itemize}

\section{RELATED WORK}
\label{chap:related_works}

\subsection{Exploration and Graph-Based Planning}

Autonomous exploration enables a robot to map an unknown environment by iteratively selecting navigation goals to maximize coverage. Early approaches addressed this problem using 2D occupancy grids~\cite{elfes2002using}, with \cite{yamauchi1997frontier} introducing the concept of frontier-based exploration. A \emph{frontier} is defined as the boundary separating known free space from unknown space; by navigating to these boundaries, the robot systematically maximizes map coverage. 
\par

While 2D representations suffice for flat indoor settings, unstructured outdoor environments require explicit modeling of terrain complexity. This necessitated the adoption of 2.5D elevation maps~\cite{fankhauser2016universal}, which capture height and slope information essential for ground robot traversability. To further handle rough terrain, risk-aware frameworks such as the STEP planner~\cite{fan2021step} augment these maps with explicit traversal cost layers, ensuring that selected exploration goals are not only geometrically reachable but also safe. However, 2.5D maps fail to accurately capture complex vertical structures like overhangs or tunnels. Consequently, scenarios requiring full spatial reasoning—most notably for aerial robots—employ dense volumetric representations like OctoMap~\cite{hornung2013octomap} and TSDF-based Voxblox~\cite{oleynikova2017voxblox} to generate frontiers in full 3D space.
\par

To improve the efficiency of coverage beyond simple frontier selection, Next-Best-View (NBV) planners~\cite{gonzalez2002navigation, bircher2016receding, isler2016information, schmid2020efficient} formulate exploration as an optimization problem. Instead of simply moving to the nearest boundary, these methods sample candidate viewpoints within a receding horizon and select the one that maximizes expected volumetric information gain by ray casting. While this produces highly efficient local paths, calculating information gain over dense volumetric maps is computationally expensive, limiting the system's ability to plan over long horizons or large areas.
\par

Scaling exploration to kilometer-scale environments requires decoupling local coverage from global spatial memory. Graph-based exploration methods address this by building a sparse topological graph of the explored environment. The Graph-Based Planner (GB-Planner)~\cite{dang2020graph} constructs a global graph where nodes represent navigable space and edges represent connectivity. This structure allows the robot to efficiently plan return paths or reposition to distant frontiers without processing the full metric map history. These graph-based architectures proved robust during the DARPA Subterranean Challenge, where teams successfully deployed them to explore kilometers of caves, tunnels, and urban underground environments~\cite {agha2021nebula, tranzatto2022cerberus}. More recent graph-based exploration planners such as FAEL~\cite{huang2023fael} and UFOExplorer~\cite{duberg2022ufoexplorer} similarly demonstrate that sparse topological structures scale far better than dense grids for kilometer-scale exploration. 
\par
Despite their success in large-scale geometric coverage, these planners remain limited by the robot's active sensor horizon. Frontiers are identified strictly based on LiDAR or depth sensor boundaries, treating all unobserved space as uniform. Consequently, these methods cannot leverage visual cues--such as a clear path visible in RGB images beyond the depth range--to prioritize promising directions. WildOS builds upon the robust topological structure of graph-based planners but incorporates a Vision Foundation Model (VFM) to score frontiers, enabling the robot to pursue promising semantic goals that lie beyond the geometric horizon.

\subsection{Open-Vocabulary Mapping and Localization}

Open-vocabulary mapping aims to ground semantic concepts in 3D space by associating vision-language features with geometric representations. Standard approaches~\cite{Peng2023OpenScene, takmaz2023openmask3d, zhang2023clip}, typically developed for indoor settings, project pixel-aligned CLIP embeddings~\cite{clip} onto dense point clouds or voxel grids. Alternatively, some approaches \cite{gu2024conceptgraphs, Werby-RSS-24} abstract these features into open-vocabulary 3D scene graphs, organizing the environment into objects, rooms, and floors. While these methods enable zero-shot queries of the environment (e.g., "find the red mug on the table") they rely on high-fidelity, pre-built maps and computationally intensive offline processing, making them unsuitable for real-time applications of field robots operating in unknown environments. \par

\medskip Recent works have attempted to extend these capabilities to handle partial observability and coarser localization. TagMap~\cite{zhang2024tagmap} decomposes the scene into a grid of voxels and identifies objects by intersecting the camera frustums of tagged views. While this provides a mechanism for coarse localization similar to triangulation, it scales poorly to large, unstructured outdoor environments due to usage of voxel-based representation. Similarly, RayFronts~\cite{alama2025rayfronts} projects vision-language features into a 3D grid, explicitly handling pixels with invalid depth by projecting them to the boundaries of the voxel map. Although it extends open-vocabulary mapping beyond the depth range, it does not estimate a coarse 3D goal position and scales poorly to large outdoor scenes. We address real-time open-vocabulary target localization in previously unseen environments by introducing a particle-based triangulation estimator that predicts coarse 3D goal locations beyond the robot’s depth sensing range, enabling goal-directed planning toward distant semantic cues.

\subsection{Object-Goal Navigation}

Navigation research has evolved from reaching specific geometric coordinates (PointGoal) to locating instances of specific semantic categories, a task formalized as Object Goal Navigation (ObjectNav)~\cite{anderson2018evaluation}. Unlike geometric navigation, ObjectNav requires the robot to explore the environment, semantically identify targets, and navigate to them without a prior map.
\par

In early works, the problem was approached from the planning side. Decision-theoretic indoor object search~\cite{kunze2012searching} formalized target discovery as planning under uncertainty over a graph of room-level locations, while \cite{hanheide2017robot} coupled symbolic task planning with explicit reasoning about open and uncertain world assumptions. Although principled, these systems relied on hand-crafted ontologies, fixed class vocabularies, and brittle perception routines that had to be hand-crafted for each new object category. Modern vision-language representations remove this engineering bottleneck by providing open-vocabulary grounding out of the box, driving the field to shift toward learned, perception-driven approaches.
\par

In indoor benchmarks, the dominant paradigm long focused on end-to-end Reinforcement Learning (RL) on simulation platforms like Habitat~\cite{savva2019habitat} and Gibson~\cite{xia2018gibson}. In these settings, agents learn policies that map raw visual observations directly to control actions~\cite{habitatchallenge2023}. While powerful in simulation, these end-to-end agents often struggle to generalize to real-world dynamics and lack the explicit safety guarantees required for field robotics. Consequently, modular approaches have gained traction, decoupling the problem into semantic mapping and geometric planning. The Semantic Exploration (SemExp) framework~\cite{chaplot2020object} builds a top-down semantic map and selects long-term goals to maximize semantic coverage. Similarly, PONI~\cite{ramakrishnan2022poni} learns a potential function to predict where objects are likely to be found based on structural cues (e.g., finding a chair near a table). ForesightNav~\cite{shah2025foresightnav} introduces an imagination strategy using a neural network that leverages current and past observations to predict unobserved regions of the environment in the form of a Bird's Eye View (BEV) map infused with CLIP~\cite{clip} features. IPPON~\cite{qu2024ippon} builds a 3D object probability map with a Bayes filter and uses LLM-derived common-sense priors to guide zero-shot open-vocabulary object-goal navigation via an RRT*-based tree of informative viewpoints. Although effective, these methods rely on learned semantic relationships between objects and structured indoor layouts--assumptions that often break down in large-scale, unstructured real-world terrains and do not readily scale to long-range exploration.

\par
Recently, Vision-Language Models (VLMs) have been employed to extend object-goal navigation beyond indoor settings. Works such as Uni-NaVid~\cite{zhang2024uni} and NaVILA~\cite{cheng2024navila} integrate VLMs with locomotion policies, enabling robots to follow open-ended instructions such as "\textit{navigate to the trash can}". However, these approaches are largely reactive; they typically lack persistent spatial memory. If the target object is not currently visible in the frame, or if the robot must search a large area without continuous explicit goal commands, these methods often fail to accomplish the task. Hierarchical planners~\cite{rana2023sayplan, Werby-RSS-24} utilize open-vocabulary semantic graphs to perform object-goal navigation, but they fundamentally rely on pre-built maps of the environment, limiting their applicability to exploration-driven object search.

\subsection{Vision-Based and Long-Range Navigation}

A prominent line of research focuses on end-to-end learning for navigation, where policies map visual observations directly to control actions~\cite{shah2021ving, shah2022gnm, shah2023vint}. These methods learn generalizable behaviors from large-scale diverse datasets, enabling robots to navigate towards image goals in novel environments. Building on this, NoMaD~\cite{sridhar2024nomad} introduces a goal-masked diffusion policy to handle exploration without explicit goal images. More recently, \cite{hirose24lelan} incorporates language conditioning into these end-to-end policies, allowing for open-ended instruction following. VLD~\cite{milikic2025vld} introduces a decoupled framework that trains a self-supervised distance predictor on internet-scale videos to guide a reinforcement learning policy, enabling scalable navigation with both image and text goals.
\par

Another line of research explicitly estimates traversability for planning. Approaches such as \cite{frey23fast, schmid2022self, jung2024v} utilize self-supervised learning to predict traversability directly in image space, associating visual features with proprioceptive feedback. In contrast, methods like \cite{meng2023terrainnet, roadrunner} estimate traversability in BEV metric maps, generating dense cost grids that better handle occlusions and enable robust multi-camera fusion. More recent works, such as SALON~\cite{sivaprakasam2025salon} and CREStE~\cite{zhang2025crestescalablemaplessnavigation}, extend this paradigm by leveraging foundation models to rapidly align these BEV maps with human preferences or to novel environments.
\par

Long-range navigation seeks to extend this planning horizon beyond the local metric map. \cite{patel2024roadrunner} addresses this by explicitly predicting traversability and elevation maps up to \SI{100}{m} by performing fusion of multiple cameras and LiDARs in BEV. LRN~\cite{schmittle2025lrn} introduces the concept of \emph{visual frontiers}--regions in the image space that appear navigable and extend beyond the visible horizon. At each timestep, the robot selects the frontier aligned best with the goal heading. However, LRN assumes implicit reachability and lacks a mechanism to assess traversability, which can lead to scoring frontiers that lie behind non-traversable obstacles. Furthermore, it has no spatial memory of previously explored or rejected frontiers, which limits robustness in ambiguous scenarios such as dead-ends. WildOS overcomes these limitations by grounding visual frontiers within a persistent navigation graph. By combining the long-range semantic insight of VFMs with the memory of a graph-based planner, our system effectively bridges the gap between far-field perception and safe, consistent global exploration.

\section{Problem Statement}
\label{sec:problem_statement}

We address the problem of \textit{long-range object search} in large-scale, unstructured outdoor environments. 
The robot, equipped with an RGB camera and a LiDAR sensor, must locate an object specified by a natural language query - $Q_{\text{goal}}$~(e.g., ``find the \textbf{water tank}''). 
The target object lies at an unknown location in a static environment $\mathcal{E}$ for which no prior map is available.

At each timestep $t$, the robot obtains a multi-modal observation 
\[
\mathcal{O}_t = \{I_t, \mathcal{T}^{\text{geo}}_t, \mathbf{x}_t\},
\]
where $I_t \in \mathbb{R}^{H\times W\times 3}$ denotes the RGB image, $\mathcal{T}^{\text{geo}}_t$ is the local geometric traversability map in the robot’s frame, derived from a 2.5D elevation grid and providing accurate traversability information within a distance of $r_{\max}$ around the robot, and $\mathbf{x}_t \in \text{SE(3)}$ represents its pose. 
The robot executes an action $a_t \in \mathcal{A}$ according to a policy $\pi : \mathcal{O}_t \mapsto a_t$, resulting in a new state $\mathbf{x}_{t+1}$.

The operator provides an approximate prior location or heading $\mathbf{g}_0$ corresponding to the text query to initiate a directed search. 
The search task concludes successfully at time T if the robot's final position $\mathbf{x}_T$ is within a predefined reachability distance $d_\textit{reach}$ of the target object specified by the textual goal $Q_{\text{goal}}$.

\paragraph{Objective.}
The goal is to find a trajectory $\xi = \{\mathbf{x}_0, \dots, \mathbf{x}_T\}$ that minimizes the expected navigation cost to reach the target object:
\begin{equation} 
\label{eq:problem_cost}
J(\xi) = \mathbb{E}_{\mathcal{E}}\!\left[ \sum_{t=0}^{T-1} C(\mathbf{x}_t, a_t, \mathbf{x}_{t+1}) \right],
\end{equation}
where $C$ denotes a task-dependent cost function incorporating travel distance, traversability, and safety constraints.
The expectation $\mathbb{E}_{\mathcal{E}}$ is taken over the distribution of possible unknown environments. The optimization is subject to sensor and computation limits that constrain planning to a local horizon $H \ll T$.
Because $J(\xi)$ cannot be measured directly on hardware, the field experiments in Sec.~\ref{sec:lrn_exps} evaluate three observable proxies: path length and mission duration reflect accumulated travel and traversability costs, while operator interventions capture the safety and recoverability constraints in $C$.

\paragraph{Challenges.}
This setting presents several challenges:
\begin{itemize}
    \item \textbf{Limited Sensing Range:} Despite LiDARs range up to \SI{150}{m} or so, their density is very low and insufficient for decision making beyond a shorter range. The local costmap $\mathcal{T}^{\text{geo}}_t$ is restricted to $r_\text{max} \approx \SI{10}{m} $, while the target object may lie several hundred meters away or occluded behind obstacles. 
    \item \textbf{Outdoor Unstructured Environments:} The environment can be characterized by hazards (e.g. fences, dense foliage) that require semantic reasoning to determine traversability beyond simple geometric models. This complexity, along with possible dead-ends in the environment, mandates long-range planning to prevent inefficient detours and backtracking.
    \item \textbf{Onboard Computation and Localization Constraints:} All perception, mapping, and planning must run in real time on the robot’s onboard computer without external communication or cloud access. In addition, global positioning signals (e.g., GNSS) are unavailable, requiring the robot to rely solely on onboard sensing and state estimation for localization during long-range navigation.
\end{itemize}

Consequently, the robot must \emph{efficiently explore and plan beyond its immediate sensor range} using visual and geometric cues to infer likely object locations and reachable paths.
Our formulation thus generalizes standard goal-directed navigation to the \emph{open-vocabulary object search} problem in long-range outdoor settings.

\section{METHODOLOGY}

\subsection{Overview}
\begin{figure*}[t]
    \centering
    \includegraphics[width=\textwidth]{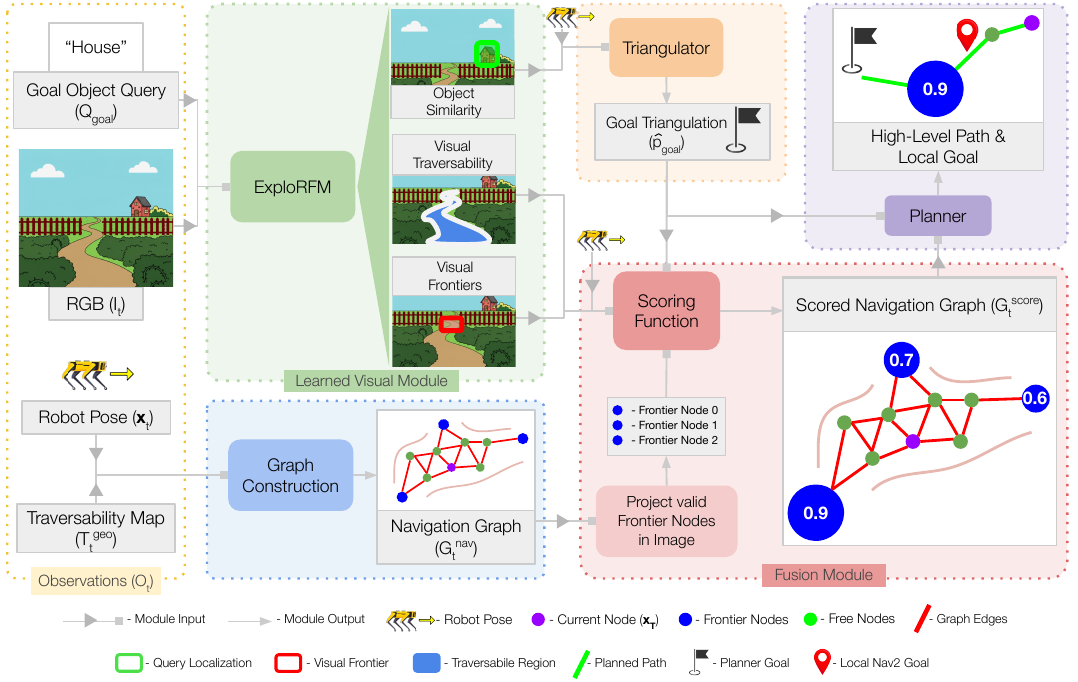}
    \caption{\textbf{Method Overview} consisting of five main components: 1) \crule[geomblue]{0.27cm}{0.27cm} WildOS incrementally builds a sparse navigation graph from geometric sensing to maintain persistent spatial memory and identify geometric frontier nodes for safe exploration (Sec.~\ref{sec:navigation_graph}). 2) \crule[visiongreen]{0.27cm}{0.27cm} To reason beyond the limited depth horizon, a learned vision-language module, ExploRFM, processes the current image and text query to predict visual traversability, visual frontiers, and open-vocabulary object similarity over a long-range visual horizon (Sec.~\ref{sec:explorfm}). 3) \crule[triangorange]{0.27cm}{0.27cm} Object detections from multiple viewpoints are fused by a probabilistic goal triangulation module to estimate a coarse 3D target location beyond direct sensor range (Sec.~\ref{sec:triangulation}). 4) \crule[fusionred]{0.27cm}{0.27cm} Geometric frontier nodes are then projected into the image and scored using the visual-semantic cues and the current goal estimate, producing a semantically scored navigation graph (Sec.~\ref{sec:scored_graph}). 5) \crule[plannerpurple]{0.27cm}{0.27cm} Finally, a hierarchical planner selects and executes actions by planning over the scored graph and generating locally safe motions toward intermediate goals (Sec.~\ref{sec:planner}).}
    \label{fig:overview}
\end{figure*}

The overall architecture of WildOS is illustrated in Fig.~\ref{fig:overview}, consisting of five main components:

\subsubsection{Navigation Graph Construction}
The local traversability map $\mathcal{T}^{\text{geo}}_t$ is incrementally integrated into a \textit{navigation graph} $G^{\text{nav}}_t = (V_t, E_t)$, where nodes $v_i \in V_t$ represent reachable locations, and edges $(v_i, v_j) \in E_t$ encode traversability costs.
Nodes lying at the boundary between explored and unexplored regions are identified as \textit{frontier nodes} $\mathcal{F}_t^{\text{geo}} \subset V_t$, which serve as candidates for exploration.  
This graph provides a memory-rich, geometry-based representation of visited regions but is limited by the range of $\mathcal{T}^{\text{geo}}_t$ i.e. $r_{\max}$, resulting in locally optimal but myopic exploration behavior.

\subsubsection{Learned Visual Module (``\textbf{ExploRFM}'')}
To reason beyond $r_{\max}$ and incorporate semantic understanding, we introduce a vision-language model named \textbf{ExploRFM} — the \textit{Exploration and Object Reasoning Foundation Model}.  
Given the current RGB frame $I_t$ and the text query $Q_{\text{goal}}$ describing the target object, ExploRFM produces three dense prediction maps:
\[
\begin{aligned}
\mathcal{T}^{\text{vis}}_t &\in [0,1]^{H \times W}, &&\text{(visual traversability map)} \\
\mathcal{F}^{\text{vis}}_t &\in [0,1]^{H \times W}, &&\text{(visual frontier score map)} \\
\mathcal{S}^{\text{vis}}_t &\in \{0,1\}^{H \times W}, &&\text{(object similarity mask)}.
\end{aligned}
\]
Here, $\mathcal{T}^{\text{vis}}_t$ estimates the semantic traversability of each pixel distinguishing between safe (e.g. grass, road) and unsafe (e.g. water, bushes) regions, $\mathcal{F}^{\text{vis}}_t$ highlights semantic regions likely to lead to novel observations or unexplored spaces, and $\mathcal{S}^{\text{vis}}_t$ localizes the region in the image corresponding to the object described by $Q_{\text{goal}}$.
While these maps provide semantic understanding over a much larger spatial range than $r_{\max}$, they lack the geometric consistency and safety guarantees of the navigation graph that, by construction, only contains edges validated against the local geometric traversability map.

\subsubsection{Goal Triangulation Module}
Detections in $\mathcal{S}^{\text{vis}}_t$ from multiple viewpoints $\{I_0, \ldots, I_t\}$ are used to triangulate a coarse 3D estimate of the goal position $\hat{\mathbf{p}}_{\text{goal}}$ by weighting sampled 3D hypotheses against the viewing rays of all detections.  This coarse localization provides a soft prior over the object’s position even when it lies beyond direct sensor range.

\subsubsection{Fusion Module: Cross-Modal Frontier Scoring}
To combine the complementary strengths of the two modalities - geometry and vision, we project each geometric frontier node $v_i \in \mathcal{F}_t^{\text{geo}}$ from world coordinates into the image plane of $I_t$.  
For nodes with valid projections, we assign a visual-semantic exploration score
\begin{equation}
\label{eq:cross_modal_score}
s_i = f_\text{score}\!\left(\mathcal{T}^{\text{vis}}_t,\, \mathcal{F}^{\text{vis}}_t,\, \pi(\mathbf{x}_{v_i}),\, \hat{\mathbf{p}}_{\text{goal}},\, \mathbf{x}_t \right),
\end{equation}
where $\pi(\cdot)$ denotes the camera projection.  
This yields a \textit{scored navigation graph} $G^{\text{score}}_t$, where each frontier node is annotated with a semantic exploration likelihood derived from vision and is conditioned on the goal estimate $\hat{\mathbf{p}}_{\text{goal}}$.

\subsubsection{Planning and Control}
A high-level planner plans a path using the scored navigation graph $G^{\text{score}}_t$ to the coarse goal estimate $\hat{\mathbf{p}}_{\text{goal}}$ (If the target object has not yet been observed, the planner falls back to the approximate prior goal $\mathbf{g}_0$ provided at mission start.) An intermediate short-range goal $\mathbf{g}^\text{local}_t$ on the path is passed to a local planner that computes a safe, dynamically feasible path and outputs control commands to the robot.  

This hierarchical design allows the robot to combine short-range geometric safety with long-range semantic reasoning, enabling efficient, vision-informed exploration and object search in large outdoor environments.

The main novelty of our work is the architecture of how these components work synergistically to enable long range visual navigation with frontier-based geometric graph memory and goal-directed search.
Next, we detail each of these components. 

\subsection{Navigation Graph}
\label{sec:navigation_graph}
\begin{figure*}[htbp]
    \centering
    \includegraphics[width=\textwidth]{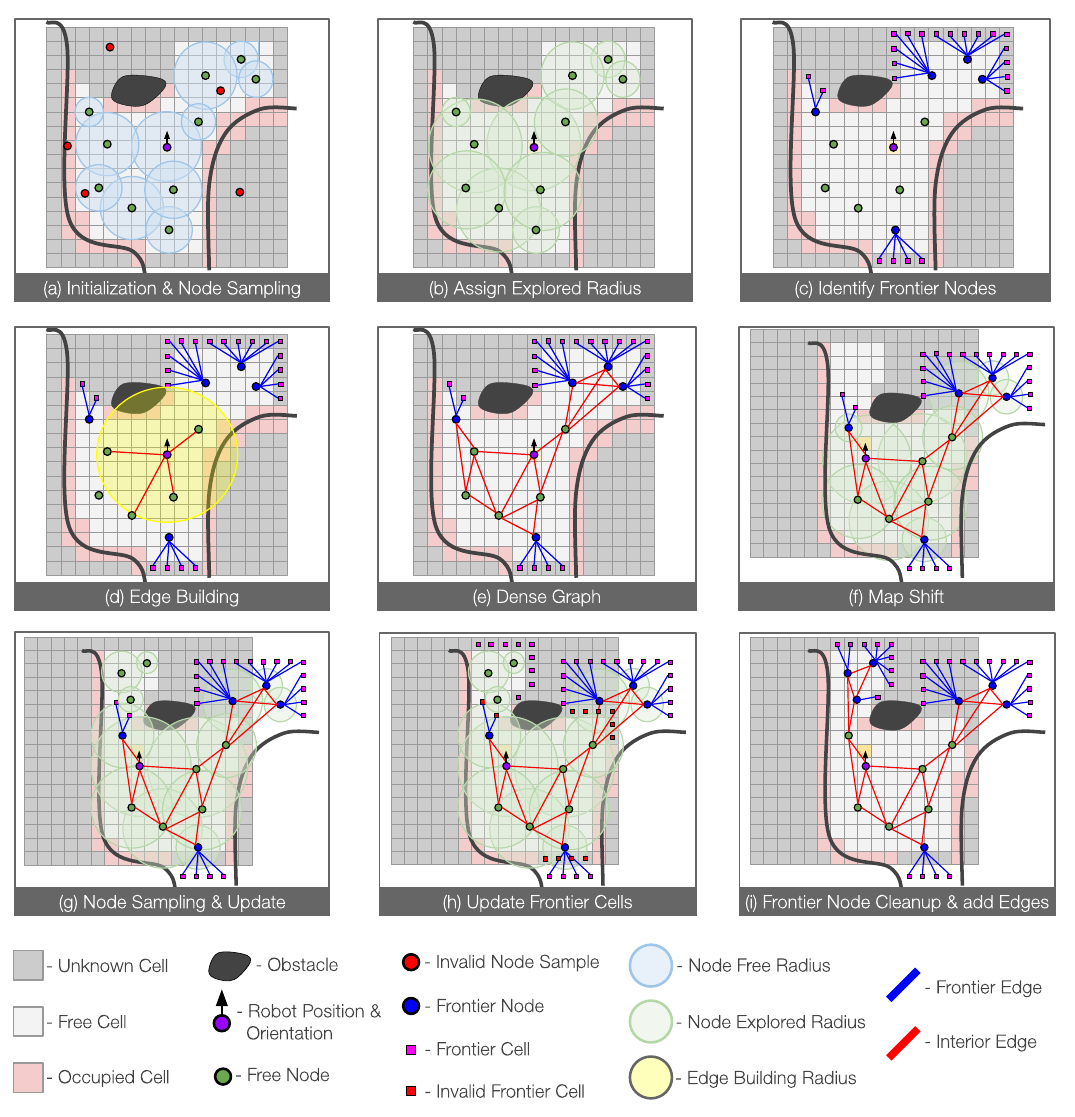}
    \caption{\textbf{Navigation Graph Construction}
    (a) Node Sampling in free cells of $\mathcal{T}^{\text{geo}}_t$, and assigning $r^f_i \text{ (free radius)}$. Invalid samples are shown in red. 
    (b) $r^e_i \text{ (explored radius)}$ for each node. 
    (c) Identifying frontier cells and assigning them to nodes to get $\mathcal{F}^{\text{geo}}_t$. 
    (d) $r_{\text{edge}}$ for the current node. 
    (e) Dense graph obtained after first iteration of graph construction. 
    (f) Robot pose $\mathbf{x}_t$ update and shift of $\mathcal{T}^{\text{geo}}_t$. 
    (g) Node Sampling and update $r^f_i, r^e_i$ 
    (h) Detect new frontier cells and remove invalid ones. Invalid ones shown in red - present in known regions, present within explored radius of another node. 
    (i) Assign frontier cells to nodes, and update frontier nodes.}
    \label{fig:nav_graph}
\end{figure*}

\subsubsection{Motivation}
Dense voxel-based maps are computationally expensive to maintain in large outdoor environments. To achieve scalable, memory-efficient exploration, we represent free-space connectivity as a sparse graph where nodes correspond to reachable regions and edges encode traversability relationships.

\subsubsection{Graph Definition}
At time $t$, the navigation graph is
\[
G^{\text{nav}}_t = (V_t, E_t),
\]
where $v_i \in V_t$ represents a collision-free location in the environment and $(v_i, v_j) \in E_t$ encodes a collision-free traversable edge between two nodes. Each node $v_i$ additionally stores:
\[
r^f_i \text{ (free radius)}, \qquad r^e_i \text{ (explored radius)},
\]
updated using signed distance fields (SDFs) to obstacles and unknown regions. 
The \emph{free radius} $r^f_i$ denotes the distance from $v_i$ to the nearest obstacle or unknown cell. It is capped by a constant $r^f_{\max}$.  
The \emph{explored radius} $r^e_i$ captures how far the environment around $v_i$ has been observed ensuring persistent memory of previously explored regions.

Nodes adjacent to unknown regions form the geometric frontier set:
\[
\mathcal{F}^{\text{geo}}_t = \{\,v_i \in V_t \mid v_i.\text{isFrontier} = \textit{true}\,\}.
\]

\subsubsection{Graph Update Procedure}
The graph is updated incrementally at each timestep using the current local traversability map $\mathcal{T}^{\text{geo}}_t$ as shown in Algorithm~\ref{alg:nav_graph_build}, and summarized in Figure~\ref{fig:nav_graph}. 
The main components of an update step include: 
\begin{enumerate}
    \item Update the free and explored radii of each node (Algorithm~\ref{alg:update_nodes})
    \item Sample new nodes in the free cells of the local map ensuring they are not at redundant locations (Algorithm~\ref{alg:sample_nodes})
    \item Detect frontier cells in the local map and update nodes near the frontier cells to frontier nodes. (Algorithm~\ref{alg:frontiers_update})
    \item Densify the graph by connecting all the nodes using collision-free edges (Algorithm~\ref{alg:edges})
\end{enumerate}

This graph-based formulation enables long-term memory and scalable mapping critical for long-range navigation. 
By encoding both geometric traversability and exploration history in a sparse structure, $G^{\text{nav}}_t$ provides a compact yet informative abstraction of the environment, enabling efficient integration with the semantic scoring and planning layers that follow.

\begin{algorithm}[ht]
\caption{Update Navigation Graph}
\label{alg:nav_graph_build}
\begin{algorithmic}[1]
\Function{UpdateNavigationGraph}{$G^{\text{nav}}_{t-1}, \mathcal{T}^{\text{geo}}_t$}
    \State $\text{SDF}^{\text{unk}}_t, \text{SDF}^{\text{obs}}_t \gets$ \Call{ComputeSDF}{$\mathcal{T}^{\text{geo}}_t$}
    \State $G^{\text{nav}}_{t-1} \gets$ \Call{UpdateNodes}{$G^{\text{nav}}_{t-1}$}
    \State $V_{\text{new}} \gets$ \Call{SampleNewNodes}{$\mathcal{T}^{\text{geo}}_t, G^{\text{nav}}_{t-1}, N_{\text{samples}}$}
    \State $V_t \gets V_{t-1} \cup V_{\text{new}}$
    \State $V_t \gets$ \Call{UpdateFrontierNodes}{$\mathcal{T}^{\text{geo}}_t, V_t$}
    \State $E^{\text{new}}_t \gets$ \Call{BuildEdges}{$V_t, \mathcal{T}^{\text{geo}}_t$}
    \State $E_t \gets E_{t-1} \cup E^{\text{new}}_t$
    \State \Return $G^{\text{nav}}_t = (V_t, E_t)$
\EndFunction
\end{algorithmic}
\end{algorithm}

\begin{algorithm}[ht]
\caption{Update Nodes}
\label{alg:update_nodes}
\begin{algorithmic}[1]
\Function{UpdateNodes}{$G^{\text{nav}}_{t-1}$}
    \ForAll{$v_i \in V_{t-1}$ within map bounds}
        \State $r^f_i \gets \min(\text{SDF}^{\text{obs}}_t(v_i),\, \text{SDF}^{\text{unk}}_t(v_i),\, r^f_{\max})$
        \State $r^e_i \gets \max(r^e_i,\, \text{SDF}^{\text{unk}}_t(v_i))$
        \If{$r^f_i = 0$} \Comment{Node lies inside obstacle}
            \State Remove $v_i$ and associated edges from $G^{\text{nav}}_{t-1}$
        \EndIf
    \EndFor
    \State \Return $G^{\text{nav}}_{t-1}$
\EndFunction
\end{algorithmic}
\end{algorithm}

\begin{algorithm}[ht]
\caption{Sample New Nodes}
\label{alg:sample_nodes}
\begin{algorithmic}[1]
\Function{SampleNewNodes}{$\mathcal{T}^{\text{geo}}_t, G^{\text{nav}}_{t-1}, N_{\text{samples}}$}
    \State $V_{\text{new}} \gets \emptyset$
    \For{$k = 1$ to $N_{\text{samples}}$}
        \State $\mathbf{p}_{\text{rand}} \sim \text{Uniform}(\text{FreeRegion}(\mathcal{T}^{\text{geo}}_t))$
        \If{$\min(\text{SDF}^{\text{obs}}_t(\mathbf{p}_{\text{rand}}), \text{SDF}^{\text{unk}}_t(\mathbf{p}_{\text{rand}})) > r_{\text{trav}}$}
            \If{$\forall v_j \in V_{t-1}, \ \|\mathbf{p}_{\text{rand}} - \mathbf{p}_{v_j}\| > r^f_j$}
                \State $V_{\text{new}} \gets V_{\text{new}} \cup \{v_{\text{rand}}\}$
            \EndIf
        \EndIf
    \EndFor
    \State \Return $V_{\text{new}}$
\EndFunction
\end{algorithmic}
\end{algorithm}

\begin{algorithm}[t]
\caption{Detect and Update Frontier Nodes}
\label{alg:frontiers_update}
\begin{algorithmic}[1]
\Function{UpdateFrontierNodes}{$\mathcal{T}^{\text{geo}}_t, V_t$}
    \LComment{Step 1: Cleanup of invalidated frontier points}
    \ForAll{$v_i \in V_t \ \textbf{with} \ v_i.\text{isFrontier} = \textbf{true}$}
        \State $v_i.\text{frontierPoints}_{\text{new}} \gets \emptyset$
        \ForAll{$p_f \in v_i.\text{frontierPoints}$}
            \If{$p_f$ is \textbf{not} in Known region of $\mathcal{T}^{\text{geo}}_t$ \textbf{and} within map bounds}
                \State Add $p_f$ to $v_i.\text{frontierPoints}_{\text{new}}$
            \EndIf
        \EndFor
        \State $v_i.\text{frontierPoints} \gets v_i.\text{frontierPoints}_{\text{new}}$
    \EndFor

    \LComment{Step 2: Update node frontier status}
    \ForAll{$v_i \in V_t$}
        \If{$v_i.\text{frontierPoints}$ is $\emptyset$}
            \State $v_i.\text{isFrontier} \gets \textbf{false}$
        \Else
            \State $v_i.\text{isFrontier} \gets \textbf{true}$
        \EndIf
    \EndFor

    \LComment{Step 3: Detect and assign new frontier points}
    \State $\mathcal{F}_{\text{cells}} \gets$ \Call{BoundaryBetween}{Free, Unknown} in $\mathcal{T}^{\text{geo}}_t$
    \ForAll{$c_f \in \mathcal{F}_{\text{cells}}$}
        \State $\mathbf{p}_{c_f} \gets$ center position of $c_f$
        \LComment{Skip if within explored radius of any node}
        \If{$\exists\, v_j \in V_t : \|\mathbf{p}_{v_j} - \mathbf{p}_{c_f}\| \le r^e_j$}
            \State \textbf{continue}
        \EndIf
        \LComment{Find nearest collision-free node}
        \State $v^* \gets \arg\min_{v_i \in V_t,\, \text{CollisionFree}(v_i, c_f)} \|\mathbf{p}_{v_i} - \mathbf{p}_{c_f}\|$
        \If{$v^*$ exists}
            \State Add $\mathbf{p}_{c_f}$ to $v^*.\text{frontierPoints}$
            \State $v^*.\text{isFrontier} \gets \textbf{true}$
        \EndIf
    \EndFor

    \State \Return $V_t$
\EndFunction
\end{algorithmic}
\end{algorithm}

\begin{algorithm}[ht]
\caption{Build Edges}
\label{alg:edges}
\begin{algorithmic}[1]
\Function{BuildEdges}{$V_t, \mathcal{T}^{\text{geo}}_t$}
    \State $E^{\text{new}}_t \gets \emptyset$
    \ForAll{$v_i \in V_t$}
        \State $\mathcal{N}_i \gets$ \Call{GetSpatialNeighbors}{$v_i, V_t, r_{\text{edge}}$}
        \ForAll{$v_j \in \mathcal{N}_i$}
            \If{CollisionFree($v_i, v_j$)}
                \State $E^{\text{new}}_t \gets E^{\text{new}}_t \cup \{(v_i, v_j)\}$
            \EndIf
        \EndFor
    \EndFor
    \State \Return $E^{\text{new}}_t$
\EndFunction
\end{algorithmic}
\end{algorithm}

\subsection{ExploRFM}
\label{sec:explorfm}
\begin{figure*}[t]
    \centering
    \includegraphics[width=\textwidth]{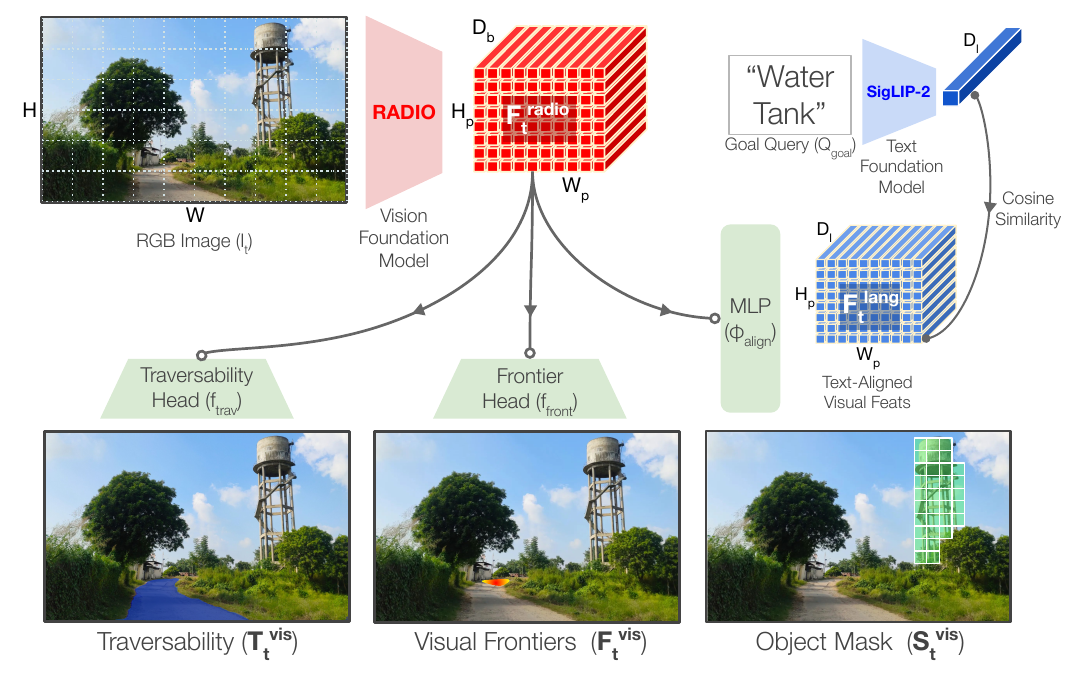}
    \caption{\textbf{ExploRFM Architecture.} ExploRFM builds upon the RADIO vision foundation model to jointly reason about traversability, semantic frontiers, and goal-object localization from a single RGB frame and language query.}
    \label{fig:explorfm}
\end{figure*}

We design \textbf{ExploRFM} (\textit{Exploration and Object Reasoning Foundation Model}) to provide long-range semantic perception that complements LiDAR-based geometry.  
Given the current RGB frame $I_t$ and a goal text query $Q_{\text{goal}}$, ExploRFM predicts three dense outputs — a visual traversability map ($\mathcal{T}^{\text{vis}}_t$), a visual frontier score map ($\mathcal{F}^{\text{vis}}_t$), and an object-similarity mask ($\mathcal{S}^{\text{vis}}_t$), as illustrated in Figure \ref{fig:explorfm}.

 $\mathcal{T}^{\text{vis}}_t \in [0,1]^{H \times W}$ provides a semantic “safety score” for each image pixel (higher the score, safer the terrain), estimating how safe it is for the robot to traverse that region based on scene semantics. Unlike LiDAR-based traversability, which is strictly geometric, this prediction accounts for visual appearance — e.g., distinguishing between \textit{grass}, \textit{dirt}, or \textit{water} — and allows reasoning about safe terrain even beyond the LiDAR’s effective range.

$\mathcal{F}^{\text{vis}}_t \in [0,1]^{H \times W}$, inspired by LRN~\cite{schmittle2025lrn}, localizes regions in the image that correspond to candidate locations for further exploration — such as the end of a trail, an opening between trees, or a road turning at a curve (higher value indicating a greater probability of the pixel being a visual frontier). We observe that such frontiers consistently emerge at the horizon of traversable regions and can be formally understood as regions in the image space lying at the boundary of the traversable contour that also align with candidate expert trajectories. Typically, these are areas at the traversable boundary which have depth discontinuities  greater than the robot size. In practice, these cues help the robot identify promising semantic directions for exploration using the prior knowledge embedded in large vision models.

\subsubsection{Backbone}
ExploRFM adopts the RADIO~\cite{radio} vision foundation model as its core. RADIO is a ViT-based model that distills features from three complementary pretrained backbones:  
DINO~\cite{oquab2023dinov2} (geometry and visual texture), CLIP~\cite{clip} (vision–language alignment), and SAM~\cite{sam} (boundary and segmentation cues).  
Given an RGB frame $I_t$, RADIO produces a dense patch-level feature map
\[
\mathbf{F}^{\text{radio}}_t \in \mathbb{R}^{H_p \times W_p \times D_b},
\quad \text{where } H_p = \tfrac{H}{p}, \ W_p = \tfrac{W}{p},
\]
with $p$ as the patch size and $D_b$ the feature dimension.

\subsubsection{Language-Aligned Feature Space}
To allow reasoning conditioned on text queries, we follow RayFronts~\cite{alama2025rayfronts} to make RADIO’s spatial features language-aligned.  
Specifically, we:  
(1) replace the final self-attention layer with the NACLIP~\cite{hajimiri2025naclip} variant using $K K^\top$ to enhance local spatial consistency, and  
(2) use RADIO’s SIGLIP summary adaptor to project the patch embeddings into a language-aligned space:  
\[
\mathbf{F}^{\text{lang}}_t = \phi_{\text{align}}(\mathbf{F}^{\text{radio}}_t) \in \mathbb{R}^{H_p \times W_p \times D_l},
\]
where $\phi_{\text{align}}$ is the pretrained MLP adaptor and $D_l$ the language-aligned feature dimension.  
For object localization, the text query $Q_{\text{goal}}$ is encoded via the Siglip-2~\cite{tschannen2025siglip} text encoder:
\[
\mathbf{q}_{\text{goal}} = \psi_{\text{text}}(Q_{\text{goal}}) \in \mathbb{R}^{D_l}.
\]
The object similarity map is computed via cosine similarity:
\[
\mathcal{S}^{\text{vis}}_t(u,v) =
\begin{cases}
1, & \text{if } \cos(\mathbf{F}^{\text{lang}}_t(u,v), \mathbf{q}_{\text{goal}}) > \tau_{\text{sim}},\\
0, & \text{otherwise},
\end{cases}
\]
and upsampled to the image resolution $(H, W)$.

\subsubsection{Traversability and Frontier Heads}
Since $\mathbf{F}^{\text{radio}}_t$ is at patch resolution, we use two independent deconvolutional CNN heads to upsample and predict per-pixel maps:
\[
\begin{aligned}
\mathcal{T}^{\text{vis}}_t &= f_{\text{trav}}\big(\mathbf{F}^{\text{radio}}_t\big),\\
\mathcal{F}^{\text{vis}}_t &= f_{\text{front}}\big(\mathbf{F}^{\text{radio}}_t\big),
\end{aligned}
\]
We discuss the architecture of the CNN heads and training details in Sec.~\ref{sec:training_explorfm}. 

\subsection{Coarse Goal Query Localization}
\label{sec:triangulation}
\begin{figure}[htbp]
    \centering
    \includegraphics[width=0.5\textwidth]{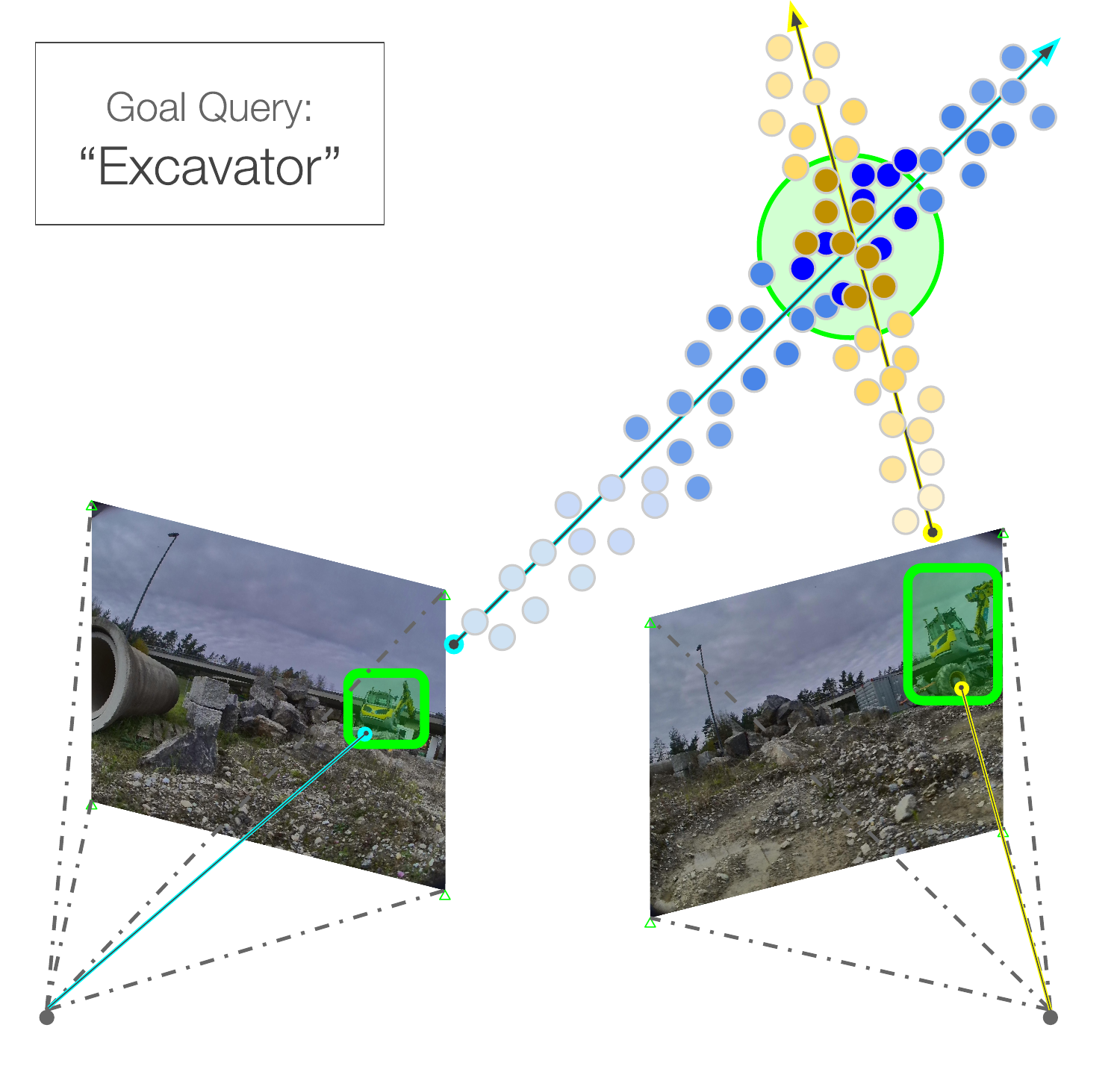}
    \caption{\textbf{Coarse goal query localization}. Ray-distance weighting based triangulation. Particles with darker shades get a higher weight.}
    \label{fig:triang-ray}
\end{figure}

While an object that falls within the LiDAR depth range can be trivially localized by taking the median depth of LiDAR points within the object mask $\mathcal{S}^{\text{vis}}_t$, our objective is to estimate a \emph{coarse} 3D location of the queried object $Q_{\text{goal}}$ even when it lies beyond LiDAR visibility.  
This enables early, directed exploration toward the target.

We propose a lightweight triangulation framework inspired by voxel-based ray intersection~\cite{alama2025rayfronts} and camera-frustum voting~\cite{zhang2024tagmap}, adapted to outdoor-scale environments through a particle-based approximation.  
Given multiple camera poses where the queried object is detected, we generate and fuse probabilistic 3D hypotheses to infer a coarse goal location.

\subsubsection{Notation and Particle Projection}
Let $\mathcal{V} = \{(\mathcal{S}^{\text{vis}}_i, \mathbf{T}^{\text{cam}}_i)\}_{i=1}^{N}$ denote the set of valid views where the object mask $\mathcal{S}^{\text{vis}}_i$ has nonzero area.  
Here, $\mathbf{T}^{\text{cam}}_i = [\mathbf{R}_i \,|\, \mathbf{t}_i] \in SE(3)$ is the camera pose at time $i$, and $\mathbf{K}$ denotes the camera intrinsics.  
The goal is to estimate $\hat{\mathbf{p}}_{\text{goal}} \in \mathbb{R}^3$ from these multi-view observations.

We first generate 3D particle hypotheses corresponding to pixels within the object masks using Algorithm \ref{alg:project_particles}.

\begin{algorithm}[H]
\caption{ProjectParticles$(\mathcal{V}, N_p, d_{\min}, d_{\max})$}
\label{alg:project_particles}
\begin{algorithmic}[1]
\State $\mathcal{P} \leftarrow \emptyset$
\For{each $(\mathcal{S}^{\text{vis}}_i, \mathbf{T}^{\text{cam}}_i) \in \mathcal{V}$}
    \State Sample $N_p$ random pixels $(u,v)$ from $\mathcal{S}^{\text{vis}}_i$
    \For{each sampled pixel $(u,v)$}
        \State Sample depth $d \sim \mathcal{U}(d_{\min}, d_{\max})$
        \State Compute 3D point in camera frame: $\mathbf{p}^{\text{cam}} = d \, \mathbf{K}^{-1} [u, v, 1]^\top$
        \State Transform to world frame: $\mathbf{p}^{\text{world}} = \mathbf{R}_i \mathbf{p}^{\text{cam}} + \mathbf{t}_i$
        \State $\mathcal{P} \leftarrow \mathcal{P} \cup \{\mathbf{p}^{\text{world}}\}$
    \EndFor
\EndFor
\State \Return $\mathcal{P}$  \Comment{Set of 3D particle hypotheses}
\end{algorithmic}
\end{algorithm}

Each particle $\mathbf{p}_k \in \mathcal{P}$ represents a plausible 3D location of the queried object based on multi-view appearance.

\subsubsection{Triangulation}
We compute the coarse object location by weighting each particle according to its alignment with the principal rays from the cameras, using Algorithm \ref{alg:ray_triang} and as illustrated in Figure \ref{fig:triang-ray}. 

\begin{algorithm}[H]
\caption{RayWeightedTriangulation$(\mathcal{P}, \mathcal{V})$}
\label{alg:ray_triang}
\begin{algorithmic}[1]
\State Initialize weights $w_k = 0$ for all $\mathbf{p}_k \in \mathcal{P}$
\For{each $\mathbf{p}_k \in \mathcal{P}$}
    \For{each view $(\mathcal{S}^{\text{vis}}_i, \mathbf{T}^{\text{cam}}_i)$}
        \State Let $\mathbf{r}_i$ be the ray from camera center to mask centroid
        \State Compute shortest distance $d_{ik}$ between $\mathbf{p}_k$ and $\mathbf{r}_i$
        \State $w_k \mathrel{+}= \frac{1}{d_{ik} + \varepsilon}$
    \EndFor
\EndFor
\State Estimate coarse location:
\[
\hat{\mathbf{p}}_{\text{goal}} = \frac{\sum_k w_k \mathbf{p}_k}{\sum_k w_k}
\]
\State \Return $\hat{\mathbf{p}}_{\text{goal}}$
\end{algorithmic}
\end{algorithm}

This formulation naturally favors particles consistent with the viewing rays of multiple observations and mitigates spurious detections from false-positive masks.
The resulting $\hat{\mathbf{p}}_{\text{goal}}$ serves as the goal for the planner until the object appears within LiDAR range for precise localization.
The scaling of the accuracy of this estimate with target distance, viewpoint baseline, mask noise, and particle budget is analyzed in Appendix~\ref{app:triangulation}.

\subsection{Constructing the Scored Navigation Graph}
\label{sec:scored_graph}
\begin{figure*}[htbp]
    \centering
    \includegraphics[width=\textwidth]{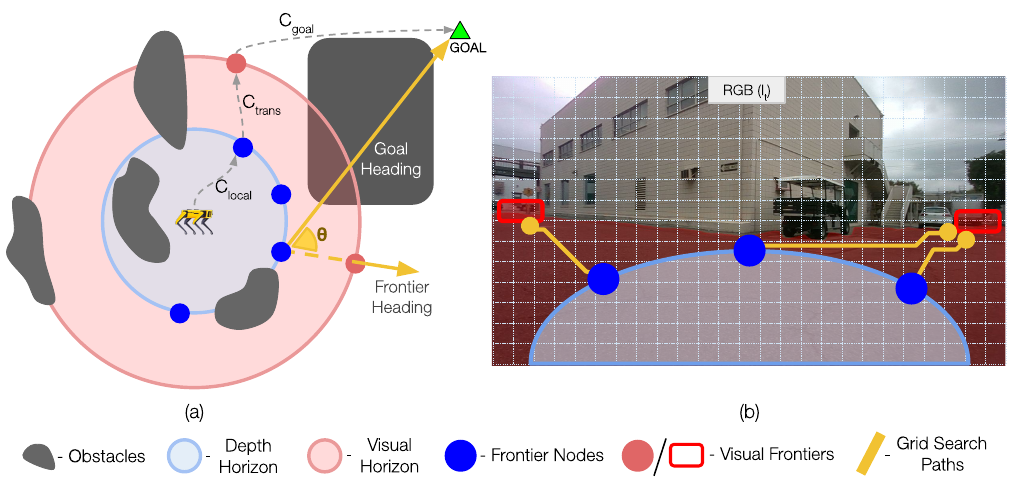}
    \caption{\textbf{Scoring the navigation graph using images.} (a) Heuristic Formulation for Distance-to-goal, and heading similarity for approximation of visual frontier to goal cost ($C_\text{goal}$) (b) In-image grid search in the traversable region for approximation of geometric frontier to visual frontier cost ($C_\text{trans}$)}
    \label{fig:scoring_navgraph}
\end{figure*}

We construct a scored navigation graph $G^{\text{score}}_t$ by integrating geometric frontiers with visual-semantic cues obtained from ExploRFM.  
At time $t$, the scored navigation graph is defined as
\[
G^{\text{score}}_t = (V_t, E_t, \mathcal{S}_t, \mathcal{D}_t),
\]
where $V_t$ and $E_t$ denote the sets of nodes and edges in $G^{\text{nav}}_t$. 
The functions 
\[
\mathcal{S}_t : \mathcal{F}^{\text{geo}}_t \rightarrow [0,1]
\quad \text{and} \quad
\mathcal{D}_t : \mathcal{F}^{\text{geo}}_t \rightarrow \mathbb{R}^{+}
\]
assign to each geometric frontier node $v_i \in \mathcal{F}^{\text{geo}}_t$ a visual-semantic score $\mathcal{S}_t(v_i)$ (computed via Eq.~\ref{eq:scoring_eq}) and the distance $\mathcal{D}_t(v_i)$ from the robot’s current pose $\mathbf{x}_T$ at which the score was last updated.

\subsubsection{From Global Objective to Local Frontier Selection}
We recall from Eq.~\eqref{eq:problem_cost} that the objective of the planner is to find a trajectory 
$\xi = \{\mathbf{x}_0, \dots, \mathbf{x}_T\}$ that minimizes the expected navigation cost
\[
J(\xi) = \mathbb{E}_{\mathcal{E}}\!\left[ \sum_{t=0}^{T-1} C(\mathbf{x}_t, a_t, \mathbf{x}_{t+1}) \right],
\]
where $C(\cdot)$ measures travel effort, traversability, and safety, and the expectation is taken over the unknown environment $\mathcal{E}$. 
Since planning is constrained to a limited local horizon $H \ll T$ by onboard sensing and computation, the robot must periodically choose a frontier node through which it will exit the current local map and continue exploration.

Accordingly, the global optimization can be decomposed into a sequence of local decisions. 
At time $t$, the planner selects a frontier node $v_i \in \mathcal{F}^{\text{geo}}_t$ that minimizes the expected total cost to reach the goal, composed of a locally known cost and a heuristic cost to continue beyond the local map.

\subsubsection{Depth and Visual Horizons}
To model perceptual limits, we define two horizons.
The \textit{depth horizon} $H_d$ represents the maximum range within which geometric mapping and traversability information are reliable—essentially the effective range of the elevation map $E_t$ i.e. $r_{\max}$.  
All geometric frontiers $\mathcal{F}^{\text{geo}}_t$ therefore lie within this depth horizon.  

Beyond $H_d$, the robot’s geometric understanding ceases, but its vision model still provides semantic and traversability cues over a farther range.  
We define this as the \textit{visual horizon} $H_v > H_d$, where visual frontiers $\mathcal{F}^{\text{vis}}_t$ appear at the boundary between semantically traversable and unknown regions in image space.  
The long-range exploration goal $\hat{\mathbf{p}}_{\text{goal}}$ typically lies beyond $H_v$, requiring the robot to reason across both horizons to plan effectively.

\subsubsection{Frontier Cost Formulation}
Within this framework, the cost of reaching the goal can be decomposed into two components (Figure~\ref{fig:scoring_navgraph}a):
a known \textit{local cost} $C_{\text{local}}$ that lies entirely within the geometric map, and an \textit{unknown heuristic cost} $C_{\text{heuristic}}$ beyond $H_d$.  
Traditionally, $C_{\text{heuristic}}$ is approximated by the Euclidean distance from the frontier to the goal~\cite{roadrunner}.  
Instead, we exploit the visual modality to estimate this cost more meaningfully using semantic and traversability cues available up to the visual horizon.

Specifically, the planner seeks the frontier node minimizing the expected total cost:
\begin{equation}
\label{eq:frontier_cost}
    v^{*} = 
    \argmin_{\substack{v_i \in \mathcal{F}^{\text{geo}}_t \\ f_j \in \mathcal{F}^{\text{vis}}_t}}
    \Big[
        C_{\text{local}}(\mathbf{x}_T, v_i)
        + C_{\text{trans}}(v_i, f_j)
        + C_{\text{goal}}(f_j, \hat{\mathbf{p}}_{\text{goal}})
    \Big],
\end{equation}
where $\mathbf{x}_T$ is the current robot pose, $C_{\text{local}}$ denotes the known cost to reach $v_i$ within the local traversability graph, $C_{\text{trans}}$ represents the transition cost from $v_i$ to a visual frontier $f_j$, and $C_{\text{goal}}$ is an approximate cost from $f_j$ to the final goal.

\subsubsection{Approximating Transition and Goal Costs}
Because geometric cost information is unavailable beyond $H_d$, we approximate these two terms using the robot’s visual-semantic representation:
\begin{itemize}
    \item \textbf{Transition cost} $C_{\text{trans}}(v_i, f_j)$  
    is estimated using the \emph{minimum-cost image path} (MCIP) between the projected frontier $\pi(v_i)$ and a visual frontier pixel $f_j$, constrained to pixels satisfying $\mathcal{T}^{\text{vis}}_t > \tau_{\text{trav}}$ (Figure~\ref{fig:scoring_navgraph}b).
    This MCIP reflects the shortest traversable path in image space and forms the basis of the \textit{reachability confidence} $R_{\text{conf}}$.
    \item \textbf{Goal cost} $C_{\text{goal}}(f_j, \hat{\mathbf{p}}_{\text{goal}})$  
    is approximated by the angular similarity between the goal direction (from $v_i$ to $\hat{\mathbf{p}}_{\text{goal}}$) and the direction of $f_j$, captured by the \textit{goal confidence} $G_{\text{conf}}$ (Figure~\ref{fig:scoring_navgraph}a).
\end{itemize}

\noindent These visual approximations enable a composite frontier score $s_i \in [0,1]$, where higher $s_i$ implies lower expected traversal cost.  
The resulting heuristic cost for planning beyond the local map is then expressed as:
\begin{equation}
\label{eq:planning_cost}
    \text{Cost}(\mathbf{x}_T, \hat{\mathbf{p}}_{\text{goal}}) 
    = \argmin_{\substack{v_i \in \mathcal{F}^{\text{geo}}_t}}
    \Big[
    d(\mathbf{x}_T, v_i)
    + z(s_i)\, d(v_i, \hat{\mathbf{p}}_{\text{goal}})
    \Big]
    ,
\end{equation}
where $d(\cdot,\cdot)$ denotes Euclidean distance and $z(\cdot)$ is a monotonically decreasing scaling function that downweights distant frontiers with high semantic confidence.

In essence, Eq.~\eqref{eq:frontier_cost} provides a cross-modal cost formulation that links the robot’s geometric understanding (within $H_d$) and visual reasoning (up to $H_v$), thereby motivating the design of our scored navigation graph.

\subsubsection{Defining the Visual-Semantic Scoring Function}
We treat each pixel in $\mathcal{F}^{\text{vis}}_t$ as a visual frontier candidate $f_j$. 
We define the projection of the node $\pi(v_i)$ as valid if the projection is within image bounds, is not behind the camera and if the node lies within a maximum scoring distance $d^{\text{score}}_\text{max}$ from the robot $\mathbf{x}_T$.
We define the scoring function $f_{\text{score}}$ introduced in Eq.~\eqref{eq:cross_modal_score} for each frontier node $v_i \in \mathcal{F}^{\text{geo}}_t$ as follows:
\begin{equation}
\label{eq:scoring_eq}
\begin{aligned}
f_{\text{score}}(v_i) =
\begin{cases}
    s_{\text{def}},\\
    \qquad \text{if } 
    \pi(v_i)\text{ invalid and } 
    v_i \notin \mathcal{F}^{\text{geo}}_{t-1}.\\[10pt]
    \displaystyle
    \max_{p \in \mathcal{X}}
    \!\left[
        G_{\text{conf}}(\hat{\mathbf{p}}_{\text{goal}}, v_i,p)\,
        R_{\text{conf}}(\pi(v_i),p)\,
        F_{\text{conf}}(p)
    \right],
    \\ \qquad\text{if } 
    \pi(v_i)\text{ valid and } 
    d(\mathbf{x}_T, v_i) < \mathcal{D}_{t-1}(v_i). \\[10pt]

    \mathcal{S}_{t-1}(v_i),
    \\\qquad \text{if }
    \pi(v_i)\text{ invalid or }
    d(\mathbf{x}_T, v_i) \ge \mathcal{D}_{t-1}(v_i). \\

\end{cases}
\end{aligned}
\end{equation}

where $\mathcal{X}$ is the set of image pixels $p=(p_x,p_y)$, $s_{\text{def}}$ is a default score assigned to newly discovered frontiers, and:
\begin{align*}
    0.5 &\leq G_{\text{conf}}(\hat{\mathbf{p}}_{\text{goal}}, v_i,p) \leq 1,\\
    0 &\leq R_{\text{conf}}(\pi(v_i), p) \leq 1,\\
    0 &\leq F_{\text{conf}}(p) \leq 1.
\end{align*}

\subsubsection{Confidence Components}
Each term in the product $G_{\text{conf}} R_{\text{conf}} F_{\text{conf}}$ contributes a specific modality-driven confidence measure:

\noindent\textbf{(a) Goal Confidence.}  
This measures alignment between the frontier heading $\mathbf{h}_{p}$ and the goal heading $\mathbf{h}_{\text{goal}}(v_i, \hat{\mathbf{p}}_{\text{goal}})$:
\begin{equation}
    G_{\text{conf}}(\hat{\mathbf{p}}_{\text{goal}}, v_i, p) = 
    \frac{3 + \mathbf{h}_{p} \cdot \mathbf{h}_{\text{goal}}(\hat{\mathbf{p}}_{v_i}, \hat{\mathbf{p}}_{\text{goal}})}{4},
\end{equation}
where $\mathbf{h}_{p}$ is the normalized ray through pixel $p$ computed via $K^{-1}[p_x,p_y,1]^T$, and $\mathbf{h}_{\text{goal}}(v_i)$ is the normalized vector from $v_i$ toward $\hat{\mathbf{p}}_{\text{goal}}$.  
This ensures $0.5 \leq G_{\text{conf}} \leq 1$ i.e. non-zero score for a frontier pointing away from the goal which would be required in scenarios where the optimal path must go in the opposite direction of the goal due to geometric constraints like deadends.

\noindent\textbf{(b) Reachability Confidence.}  
The reachability confidence between $\pi(v_i)$ and $p$ is, computed over the image grid graph restricted to pixels with $\mathcal{T}^{\text{vis}}_t > \tau_{\text{trav}}$:
\begin{align}
    w(p) &= \frac{1}{\mathcal{T}^{\text{vis}}_t(p) + \varepsilon},\\
    R_{\text{conf}}(\pi(v_i), p) &= 1 - \tanh\!\left(\frac{\text{MCIP}(\pi(v_i), p)}{H + W}\right),
\end{align}
where $\text{MCIP}(\pi(v_i), p)$ denotes the minimum-cost image path distance between $\pi(v_i)$ and $p$, with edge weights $w(p)$. 

\noindent\textbf{(c) Frontier Confidence.}  
This term suppresses unreliable or low-confidence frontier predictions:
\begin{equation}
    F_{\text{conf}}(p) =
    \begin{cases}
        \mathcal{F}^{\text{vis}}_t(p), & 
        \text{if } 
        \mathcal{F}^{\text{vis}}_t(p) > \tau_{\text{front}},\\[4pt]
        0, & \text{otherwise.}
    \end{cases}
\end{equation}

We provide an overview of the full scoring pipeline in Algorithm~\ref{alg:scored_graph}.

\begin{algorithm}[t]
\caption{Scored Frontier Graph Construction}
\label{alg:scored_graph}
\begin{algorithmic}[1]
\Require $G^{\text{nav}}_t$, $\mathcal{T}^{\text{vis}}_t$, $\mathcal{F}^{\text{vis}}_t$, $I_t$, $\mathbf{x}_T$, $\hat{\mathbf{p}}_{\text{goal}}$, $\mathcal{S}_{t-1}$, $\mathcal{D}_{t-1}$
\For{each $v_i \in \mathcal{F}^{\text{geo}}_t$}
    \If{$\pi(v_i)\text{ invalid and } v_i \notin \mathcal{F}^{\text{geo}}_{t-1}$}
        \State $\mathcal{S}_t(v_i) \gets s_{\text{def}}$
        \State $\mathcal{D}_t(v_i) \gets \infty$
    \ElsIf{$\pi(v_i)$ valid and (($v_i \notin \mathcal{F}^{\text{geo}}_{t-1}$) or ($d(\mathbf{x}_T, v_i) < \mathcal{D}_{t-1}(v_i)$))}
        \For{each pixel $p \in \mathcal{X}$ where $\mathcal{T}^{\text{vis}}_t(p) > \tau_{\text{trav}}$}
            \State compute $G_{\text{conf}}(\hat{\mathbf{p}}_{\text{goal}},v_i,p)$, $R_{\text{conf}}(\pi(v_i),p)$, $F_{\text{conf}}(p)$
            \State $s_i(p) \gets G_{\text{conf}}(\hat{\mathbf{p}}_{\text{goal}},v_i,p)\,R_{\text{conf}}(\pi(v_i),p)\,F_{\text{conf}}(p)$
        \EndFor
        \State $\mathcal{S}_t(v_i) \gets \max_{p \in \mathcal{X}} s_i(p)$
        \State $\mathcal{D}_t(v_i) \gets \text{dist}(\mathbf{x}_T, v_i)$
    \Else
        \State $\mathcal{S}_t(v_i) \gets \mathcal{S}_{t-1}(v_i)$
        \State $\mathcal{D}_t(v_i) \gets \mathcal{D}_{t-1}(v_i)$
    \EndIf
\EndFor
\State \Return $G^{\text{score}}_t = (V_t, E_t, \mathcal{S}_t, \mathcal{D}_t)$
\end{algorithmic}
\end{algorithm}

\subsubsection{Implementation Detail: Goal-Agnostic Scoring}
The frontier scoring function defined in Eq.~\eqref{eq:cross_modal_score}
\begin{equation*}
s_i = f_\text{score}\!\left(\mathcal{T}^{\text{vis}}_t,\, \mathcal{F}^{\text{vis}}_t,\, \pi(\mathbf{x}_{v_i}),\, \hat{\mathbf{p}}_{\text{goal}},\, \mathbf{x}_t \right),
\end{equation*}
is implemented so that it is \textit{goal-agnostic}, i.e. it does not explicitly depend on the goal estimate $\hat{\mathbf{p}}_{\text{goal}}$. 
This design choice is motivated by the fact that the goal localization module continuously updates $\hat{\mathbf{p}}_{\text{goal}}$ as new observations arrive. 
If the score $\mathcal{S}_t(v_i)$ were conditioned on a previous (and potentially outdated) goal estimate, it would become invalid once the goal position changes, requiring costly re-scoring of all frontier nodes.

To address this, we employ a \textbf{goal-agnostic multi-heading scoring} strategy.  
Instead of computing a single scalar score for a specific goal direction, we evaluate $f_\text{score}$ across a discretized set of possible goal headings around each frontier node $v_i$.  
The continuous space of possible goal heading directions (a full $360^\circ$ around $v_i$) is uniformly divided into $N_{\text{bins}}$ angular bins.  
For each bin $b_k \in \{1, 2, \ldots, N_{\text{bins}}\}$, we define a representative goal heading vector $\mathbf{h}^{(k)}_{\text{goal}}$, corresponding to the center of the angular sector.  
A distinct score is then computed for each bin:
\begin{equation}
\label{eq:multi_heading_score}
\begin{split}
s_i^{(k)} = f_\text{score}\!\left(\mathcal{T}^{\text{vis}}_t,\, \mathcal{F}^{\text{vis}}_t,\, \pi(\mathbf{x}_{v_i}),\, \mathbf{h}^{(k)}_{\text{goal}},\, \mathbf{x}_t\right), \\
\quad k = 1, \ldots, N_{\text{bins}}.
\end{split}
\end{equation}

Consequently, the score attribute of the scored navigation graph becomes a vector of floating-point values rather than a scalar:
\[
\mathcal{S}_t(v_i) = 
\big[\, s_i^{(1)},\, s_i^{(2)},\, \ldots,\, s_i^{(N_{\text{bins}})} \,\big],
\quad v_i \in \mathcal{F}^{\text{geo}}_t.
\]

\noindent The formal definition of the scored navigation graph is thus updated to:
\[
G^{\text{score}}_t = (V_t, E_t, \mathcal{S}_t, \mathcal{D}_t),
\quad 
\mathcal{S}_t : \mathcal{F}^{\text{geo}}_t \rightarrow [0,1]^{N_{\text{bins}}}.
\]

\subsection{Planner}
\label{sec:planner}
\begin{figure*}[htbp]
    \centering
    \includegraphics[width=\textwidth]{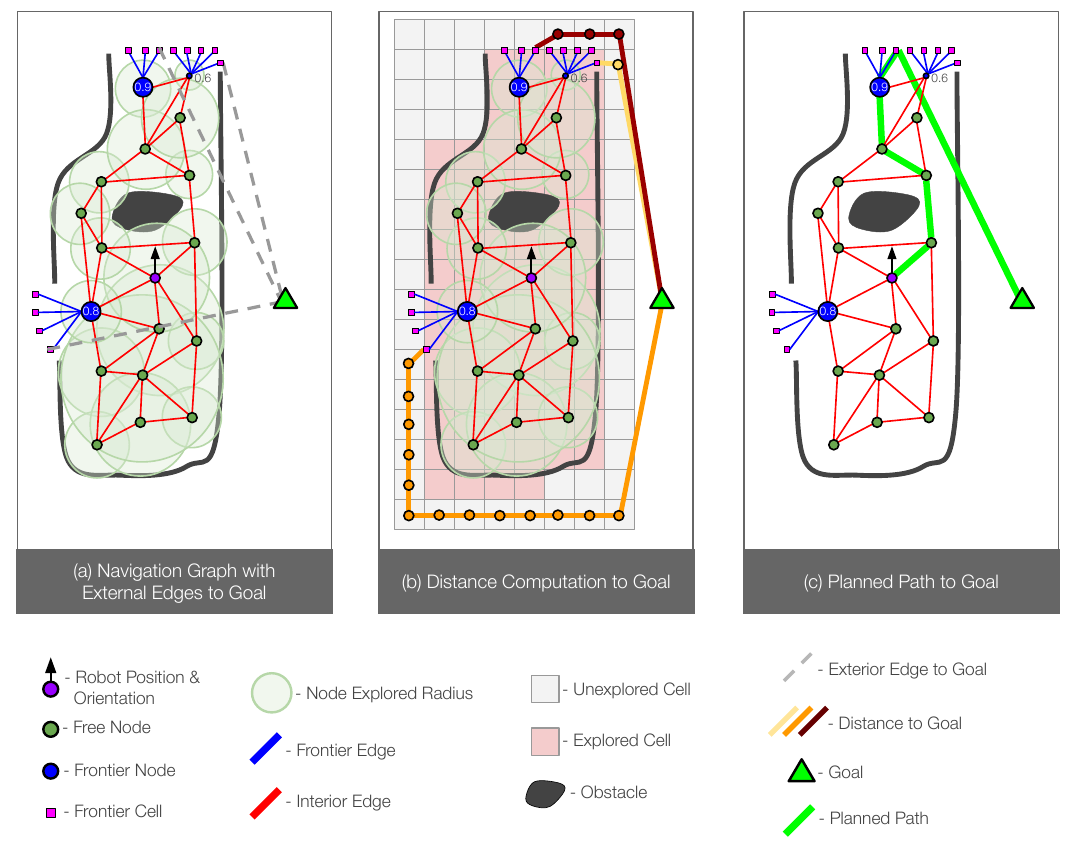}
    \caption{\textbf{Planning using the Scored Graph.}
    (a) Visualizing edges to the auxiliary goal node. (b) Computing cost of edges from frontier nodes to the goal node using shortest collision-free path search in a coarse grid. A darker color implies a higher traversal cost. (c) Final high-level path after Dijkstra's algorithm on the graph.}
    \label{fig:planning}
\end{figure*}

Given the scored navigation graph $G^{\text{score}}_t$ and a target goal position $\hat{\mathbf{p}}_{\text{goal}}$, the objective is to compute a high-level path that guides the robot towards the goal by leveraging both geometric connectivity and visual information.

\subsubsection{Cost Formulation}
We employ the cost function previously defined in Eq.~\eqref{eq:planning_cost}, where the total cost to the goal is decomposed into a geometric distance term and a visually modulated heuristic:
\begin{equation*}
    \text{Cost}(\mathbf{x}_T, \hat{\mathbf{p}}_{\text{goal}}) 
    = \argmin_{\substack{v_i \in \mathcal{F}^{\text{geo}}_t}}
    \Big[
    d(\mathbf{x}_T, v_i)
    + z(s_i)\, d(v_i, \hat{\mathbf{p}}_{\text{goal}})
    \Big]
    ,
\end{equation*}
where $d(\cdot, \cdot)$ denotes Euclidean distance, and $z(\cdot)$ is a monotonically decreasing scaling function that adjusts the contribution of the heuristic based on the visual--semantic score $s_i$.  
We empirically use 
\begin{equation}
\label{eq:planning_scale_func}
z(s_i) = 1 - \alpha \log(s_i + \epsilon),
\end{equation}
where $\alpha > 0$ and $\epsilon$ is a small constant to ensure numerical stability.

\subsubsection{Graph Construction for Planning}
Planning over $G^{\text{score}}_t$ can be reduced to a shortest-path search if we introduce an auxiliary goal node $\hat{v}_{\text{goal}}$ and connect each geometric frontier node $v_i \in \mathcal{F}^{\text{geo}}_t$ to $\hat{v}_{\text{goal}}$ through an additional edge $e_{\text{goal}}(v_i)$.  
The traversal cost of this edge is defined as
\[
C(e_{\text{goal}}(v_i)) = z(s_i^{(j)}) \, \tilde{d}(v_i, \hat{\mathbf{p}}_{\text{goal}}),
\]
where $s_i^{(j)}$ denotes the score of the frontier node corresponding to the goal heading bin $j$, and $\tilde{d}(v_i, \hat{\mathbf{p}}_{\text{goal}})$ is an approximation of the traversable distance from $v_i$ to the goal.

\subsubsection{Safe Distance Approximation}
The naive Euclidean distance between $v_i$ and $\hat{\mathbf{p}}_{\text{goal}}$ can underestimate the true traversal cost if the straight line passes through previously explored or occupied regions.  
To correct for this, we construct a coarse occupancy grid $\mathcal{G}^{\text{coarse}}$ around $G^{\text{score}}_t$:
\begin{itemize}
    \item Cells intersecting the explored radius of any node are marked as explored (occupied).
    \item The remaining cells are marked unexplored (free).
\end{itemize}
Edges from cells on the boundary of $\mathcal{G}^{\text{coarse}}$ to the goal are only added if they do not intersect explored cells.  
The distance $\tilde{d}(v_i, \hat{\mathbf{p}}_{\text{goal}})$ is then computed as the shortest collision-free path length in this coarse grid, ensuring the goal cost is never under-approximated (see Figure~\ref{fig:planning}).

\subsubsection{Graph Search}
With these augmented edges, standard graph search algorithms (e.g. Dijkstra’s or A*) can be directly applied to $G^{\text{score}}_t$.  
The start node is chosen as the node closest to the robot’s current position $\mathbf{x}_T$, and the search continues until $\hat{v}_{\text{goal}}$ is reached.  
The resulting path represents a high-level plan that traverses frontiers prioritized by their visual--semantic potential.

A short-range local goal $\mathbf{g}^\text{local}_t$ is then selected along this high-level path at a distance $d_{\text{local}}$ from the robot.  
$\mathbf{g}^\text{local}_t$ is subsequently passed to the local planner, which generates fine-grained control commands for navigation. Algorithm ~\ref{alg:planner} provides an overview of the planner.

\begin{algorithm}[H]
\caption{High-Level Planning on the Scored Navigation Graph}
\label{alg:planner}
\begin{algorithmic}[1]
\Require $G^{\text{score}}_t = (V_t, E_t, \mathcal{S}_t, \mathcal{D}_t)$, $\mathbf{x}_T$, $\hat{\mathbf{p}}_{\text{goal}}$
\Ensure High-level path $\mathcal{P}_{\text{global}}$
\State Identify start node $v_{\text{start}} = \argmin_{v_i \in V_t} d(\mathbf{x}_T, v_i)$
\ForAll{$v_i \in \mathcal{F}^{\text{geo}}_t$}
    \State Compute goal heading $\mathbf{h}_{\text{goal}}(v_i) = \frac{\hat{\mathbf{p}}_{\text{goal}} - \mathbf{x}_{v_i}}{\|\hat{\mathbf{p}}_{\text{goal}} - \mathbf{x}_{v_i}\|}$
    \State Determine heading bin $j = \text{BinIndex}(\mathbf{h}_{\text{goal}}(v_i))$
    \State Compute safe goal distance $\tilde{d}(v_i, \hat{\mathbf{p}}_{\text{goal}})$ using coarse grid search
    \State Compute goal edge cost $C(e_{\text{goal}}(v_i)) = z(\mathcal{S}_t(v_i)^{(j)})\,\tilde{d}(v_i, \hat{\mathbf{p}}_{\text{goal}})$
    \State Add edge $e_{\text{goal}}(v_i)$ connecting $v_i$ to $\hat{v}_{\text{goal}}$ with cost $C(e_{\text{goal}}(v_i))$
\EndFor
\State $\mathcal{P}_{\text{global}} \leftarrow$ Dijkstra($G^{\text{score}}_t$, $v_{\text{start}}$, $\hat{v}_{\text{goal}}$)
\State Select local goal $\mathbf{g}^\text{local}_t$ at distance $d_{\text{local}}$ along $\mathcal{P}_{\text{global}}$
\State \Return $\mathcal{P}_{\text{global}}$
\end{algorithmic}
\end{algorithm}

\section{EXPERIMENTS \& DISCUSSION}
In this section, we empirically evaluate WildOS - our proposed framework for language-conditioned object search and long-range navigation in large unstructured outdoor environments.  
Our experiments are designed to systematically answer a set of targeted research questions that evaluate the system holistically and in isolation, across multiple environments and configurations.

\medskip
We first describe the overall experimental setup, including deployment details and ExploRFM training, followed by a series of experiments aimed at answering four key questions:
\begin{itemize}
    \item \textbf{Q1:} Does the complete WildOS system enable successful end-to-end object search from language queries? (Sec.~\ref{sec:objsearch_experiments})
    \item \textbf{Q2:} Does integrating vision-based scoring with the navigation graph improve navigation performance compared to pure-geometry or pure-vision approaches? (Sec.~\ref{sec:q2}) 
    \item \textbf{Q3:} Does the navigation graph improve robustness and memory compared to purely vision-based navigation? (Sec.~\ref{sec:q3_deadend}) 
    \item \textbf{Q4:} Does WildOS generalize effectively across diverse outdoor terrains? (Sec.~\ref{sec:q4_generalization})
\end{itemize}

\subsection{System-Level Overview and Deployment Details}
\label{sec:system_overview}

\begin{figure*}[htbp]
    \centering
    \includegraphics[width=\textwidth]{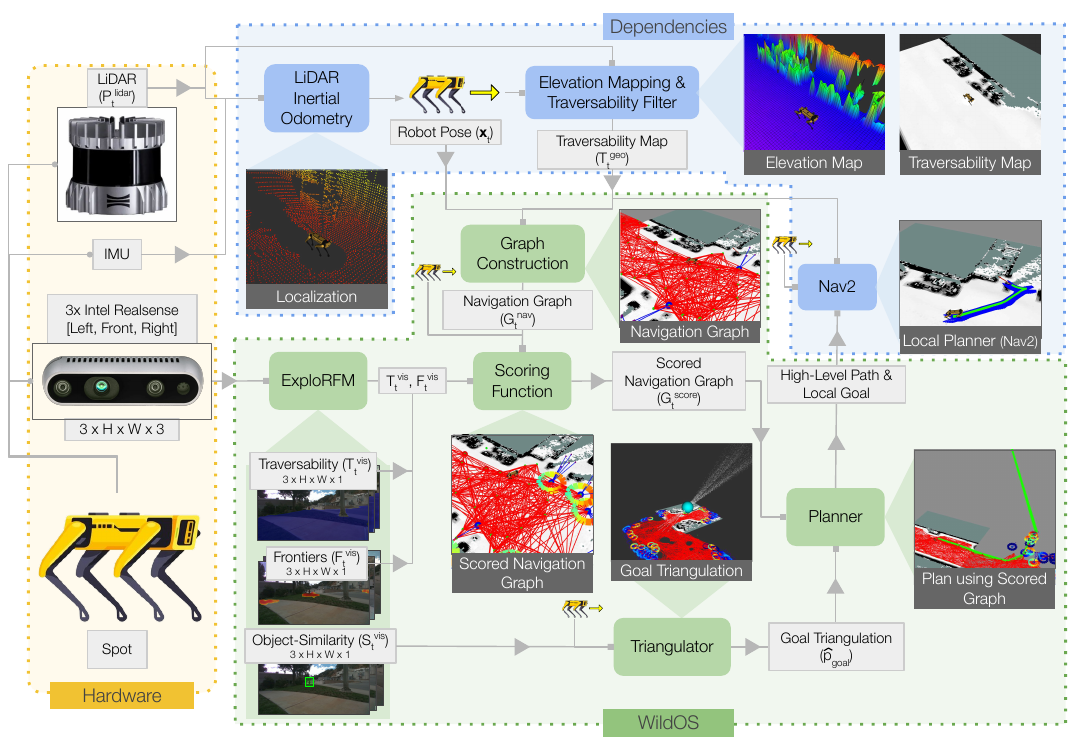}
    \caption{\textbf{System Overview}. Frontiers and Traversability are thresholded and shown in 
\crule[jetred]{0.22cm}{0.22cm} Jet and 
\crule[jetblue]{0.22cm}{0.22cm} Inverse Jet colormaps respectively. 
Edges of the navigation graph are shown in 
\crule[jetred]{0.22cm}{0.22cm} red, free nodes in 
\crule[freeGreen]{0.22cm}{0.22cm} green, and frontier nodes in 
\crule[frontierBlue]{0.22cm}{0.22cm} blue. 
Frontier nodes are surrounded with a score ring in the Scored Navigation Graph, indicating the score 
(\crule[scoreJet]{0.22cm}{0.22cm} color according to the Jet colormap) in the goal direction. 
The Triangulated Goal is shown as a 
\crule[cyanGoal]{0.22cm}{0.22cm} cyan sphere, along with projected particles in 
\crule[particleWhite]{0.22cm}{0.22cm} white. 
The planned high-level path is shown in 
\crule[pathGreen]{0.22cm}{0.22cm} green. 
A similar color scheme is followed in subsequent figures unless stated otherwise.}
    \label{fig:sys_overview}
\end{figure*}

WildOS is deployed on a Boston Dynamics \textit{Spot} quadruped platform. An overview of the full system architecture is illustrated in Fig.~\ref{fig:sys_overview}. Table~\ref{tab:system_params} lists all the key parameters, their descriptions and values used during deployment. 

\begin{table*}[t!]
\centering
\caption{\textbf{WildOS System Parameters}}
\label{tab:system_params}
\renewcommand{\arraystretch}{0.8}
\setlength{\tabcolsep}{6pt}
\begin{tabular}{p{3.2cm} p{6.4cm} p{1.2cm} p{2cm}}
\toprule
\textbf{Parameter} & \textbf{Description} & \textbf{Symbol} & \textbf{Value} \\
\midrule
\multicolumn{4}{c}{\textbf{A. Geometric Mapping \& Navigation Graph}} \\
\midrule
Local map radius & Radius of reliable geometric sensing & $r_{\max}$ & 10 m \\
Local map resolution & Resolution of $\mathcal{T}^{\text{geo}}_t$ & — & 0.1 m \\
Max free radius & Cap on node free-space radius & $r^f_{\max}$ & 4.0 m \\
Node sampling count & Node samples per update step & $N_{\text{samples}}$ & 1000 \\
Traversability clearance & Min obstacle clearance for node & $r_{\text{trav}}$ & 0.5 m \\
Edge connection radius & Max distance for graph edges & $r_{\text{edge}}$ & 8.0 m \\
\midrule
\multicolumn{4}{c}{\textbf{B. ExploRFM (Vision-Language Module)}} \\
\midrule
Image resolution & Input RGB resolution & $H \times W$ & $540 \times 960$ \\
Backbone patch size & RADIO patch size & $p$ & 16 px \\
Backbone feature dim & Dimensionality of RADIO backbone features & $D_b$ & 768 \\
Language feature dim & Dimensionality of language aligned features & $D_l$ & 1152 \\
Model Precision & -- & -- & fp16 \\
Similarity threshold & Object similarity threshold & $\tau_{\text{sim}}$ & 0.09 \\
Traversability threshold & Pixel considered traversable & $\tau_{\text{trav}}$ & 0.9 \\
Frontier threshold & Pixel considered visual frontier & $\tau_{\text{front}}$ & 0.6 \\
Model inference rate & ExploRFM inference rate (Jetson AGX Orin) & -- & 1.4 Hz \\
\midrule
\multicolumn{4}{c}{\textbf{F. Training Details (ExploRFM Heads)}} \\
\midrule
Traversability loss & Loss for $f_{\text{trav}}$ head & — & BCE \\
Frontier loss & Loss for $f_{\text{front}}$ head & — & Weighted BCE \\
Frontier positive class wt & Positive class weighting factor in Weighted BCE loss & — & 4 \\
Batch size & -- & — & 8 \\
Learning rate & -- & — & $5\times10^{-5}$ \\
Training epochs & -- & — & 100 \\
Optimizer & -- & — & Adam \\
\midrule
\multicolumn{4}{c}{\textbf{C. Goal Triangulation}} \\
\midrule
Particles per view & Samples drawn per valid detection & $N_p$ & 1000 \\
Min ray depth & Min sampled triangulation depth & $d_{\min}$ & 1 m \\
Max ray depth & Max sampled triangulation depth & $d_{\max}$ & 100 m \\
\midrule
\multicolumn{4}{c}{\textbf{D. Visual–Geometric Frontier Scoring}} \\
\midrule
Default frontier score & Score for newly seen frontier (invalid projection) & $s_{\text{def}}$ & 0.3 \\
Scoring distance limit & Max distance to consider scoring & $d^{\text{score}}_{\max}$ & 9 m \\
Heading bins & Discrete goal-heading bins & $N_{\text{bins}}$ & 16 \\
\midrule
\multicolumn{4}{c}{\textbf{E. Planning}} \\
\midrule
Planning horizon (local) & Local planner lookahead distance & $d_{\text{local}}$ & 5 m \\
Heuristic scaling factor & Weight in score-to-cost mapping & $\alpha$ & 20 \\
Goal acceptance radius & Distance threshold for declaring navigation goal reached & $d_\textit{reach}$ & 0.5 \\
\bottomrule
\end{tabular}
\end{table*}

\paragraph{Sensing and Localization.}
The robot is equipped with an Ouster OS0-128 LiDAR, a VectorNav VN-100 IMU, and three Intel RealSense D455 RGB--D cameras mounted on the left, front, and right sides of the body.  
All modules communicate through the \textsc{ROS\,2} framework.  
Robot localization is performed using \textsc{DLIO}~\cite{dlio}, a LiDAR–inertial odometry method that fuses LiDAR scans $\mathcal{P}^{\text{lidar}}_t$ with IMU measurements to produce an odometric pose estimate $\mathbf{x}_T$ relative to the start frame $\mathbf{x}_0$. 

\paragraph{Elevation and Traversability Mapping.}
The LiDAR point cloud $\mathcal{P}^{\text{lidar}}_t$ and the current pose $\mathbf{x}_T$ are processed by an elevation mapping module using the GPU-accelerated library of Elevation Mapping CuPy~\cite{miki2022elevation}.  
This node constructs a local elevation grid
\[
E_t \in \mathbb{R}^{M \times M},
\]
where each cell represents the estimated ground height relative to the robot.  
Cells without valid LiDAR returns are marked as NaN. 
A traversability filter subsequently converts $E_t$ into a traversability map
\[
\mathcal{T}^{\text{geo}}_t \in [0,1]^{M \times M},
\]
where $1$s indicate safe terrain.  
This map forms part of the observation $\mathcal{O}_t = \{I_t, \mathcal{T}^{\text{geo}}_t, \mathbf{x}_T\}$ used by WildOS as the geometric input to our navigation graph construction.

\paragraph{Visual Processing and Frontier Scoring.}
The visual processing pipeline operates using images from all three RGB cameras.  
We batch the synchronized images $\{I_t^{\text{left}}, I_t^{\text{front}}, I_t^{\text{right}}\}$ into a tensor of size $3 \times H \times W \times 3$ before processing using ExploRFM. Coarse goal localization uses object similarity $\mathcal{S}^{\text{vis}}_t$ from all views. For cross-modal scoring of frontier nodes, each geometric frontier is projected into all three image planes and scored using Eq.~\eqref{eq:multi_heading_score}. The lateral cameras help to get a baseline for the goal triangulation and cover all candidate exploration directions for the scoring.

\paragraph{Planning and Control.}
For local motion planning towards the short-range waypoint $\mathbf{g}^\text{local}_t$ given by the planner, we employ the \textsc{Nav2}~\cite{nav2} stack within \textsc{ROS\,2}, which computes dynamically feasible trajectories and generates low-level velocity commands for the Spot base controller.  
All goal-heading computations and reachability checks are performed in 2D on the ground-projected frame.

\paragraph{Onboard Computation.}
All perception and planning modules run fully onboard on two compute units.  
Localization (DLIO~\cite{dlio}) and local motion planning (Nav2~\cite{nav2}) execute on an Intel NUC\,i7, while elevation mapping, traversability estimation, and WildOS (including the neural network ExploRFM) run on an NVIDIA Jetson AGX Orin GPU. The C++ implementation of the navigation-graph maintenance (sampling, SDF updates, frontier detection, edge construction) runs at \SI{5}{Hz} on a single CPU core of the NVIDIA Jetson AGX Orin. The GPU utilization of ExploRFM is the principal bottleneck: joint inference on three synchronized cameras runs at $\sim\!\SI{1.4}{Hz}$ in fp16 ($\sim\!\SI{700}{ms}$ per call) on the C-RADIO-V3-B backbone. Scored-graph projection and frontier scoring add $\sim\!\SI{120}{ms}$, and the triangulation module itself is negligible at well under \SI{10}{ms} per call. DLIO and Nav2 run on the NUC and are unaffected by the GPU loop.

\subsection{Training ExploRFM}
\label{sec:training_explorfm}
We employ the \textbf{C-RADIO-V3-B} backbone~\cite{radio} (\SI{90}{M} parameters) as the core feature extractor for ExploRFM.
Both the traversability head ($f_{\text{trav}}$) and the frontier head ($f_{\text{front}}$) are lightweight deconvolutional decoder heads - each consisting of four blocks of \texttt{[DeConv--Conv--ReLU]} layers, followed by a sigmoid activation in the final layer to predict dense binary maps. The language head ($\phi_{\text{align}}$ - a MLP) is already pretrained and provided by RADIO~\cite{radio} authors.

\paragraph{Traversability Head.}
We supervise the traversability decoder using the \textbf{RUGD} dataset~\cite{RUGD2019IROS}, a large-scale off-road semantic segmentation dataset comprising $7436$ images spanning $24$ semantic classes.  
From the semantic annotations, we generate binary traversability masks by assigning a label of $1$ (traversable) to pixels belonging to the categories
\[
\text{safe\_labels} =
\left\{
\begin{array}{l}
\texttt{dirt, sand, grass,}\\
\texttt{asphalt, gravel, mulch,}\\
\texttt{rock-bed, concrete}
\end{array}
\right\}
\]

and $0$ (non-traversable) otherwise.  
We train the head using binary cross-entropy (BCE) loss with an $80$/$20$ train/validation split.  
This setup enables the model to learn a semantic notion of terrain safety from diverse real-world conditions.

\paragraph{Frontier Head.}
To train the frontier decoder, we manually annotate $350$ images from the \textbf{GrandTour} dataset~\cite{grandtour}, which covers a wide range of terrains including off-road, urban, industrial, mountain, and forest scenes.  
For each image, we mark bounding boxes around visually salient exploration candidates (trail ends, openings, or turns) and assign all pixels within each box as frontier ($1$), and all others as non-frontier ($0$).  
Due to the inherent class imbalance between frontier and background pixels, we employ a \textbf{weighted BCE loss}.  
The decoder shares the same lightweight architecture as the traversability head.  
Importantly, we train only this small task-specific head on top of frozen foundation-model features, keeping training efficient and data-light. Training the frontier head takes approximately $1$ hour and $26$ minutes on a single NVIDIA GeForce RTX 4090 GPU. 
Despite the small number of samples, the resulting model exhibits strong generalization, highlighting the expressiveness of the RADIO feature space.

\paragraph{Held-Out Accuracy.}
Table~\ref{tab:head_metrics} reports the accuracy of both heads on the held-out $20\%$ split of their respective datasets.  
The traversability head scores well on all three metrics, as terrain classes are visually distinct and the masks derived from the RUGD labels are unambiguous.  
The frontier head has high pixel accuracy but low mIoU and F1.  
This is expected: frontier pixels cover only a small part of the image, and a scene typically contains several valid openings of which the annotator boxed only some, so correct predictions are often counted as false positives.  
The low overlap scores matter little in practice, since $\mathcal{F}^{\text{vis}}_t$ enters Eq.~\eqref{eq:scoring_eq} through a $\max$ over pixels: the planner uses where the strongest activations are, not how large they are.

\begin{table}[htbp]
\centering
\caption{\textbf{Held-out accuracy of the ExploRFM heads}, measured on the $20\%$ validation split of RUGD~\cite{RUGD2019IROS} for $f_{\text{trav}}$ and of the annotated GrandTour~\cite{grandtour} subset for $f_{\text{front}}$.}
\label{tab:head_metrics}
\renewcommand{\arraystretch}{1.2}
\setlength{\tabcolsep}{6pt}
\begin{tabular}{lccc}
\toprule
\textbf{Head} & \textbf{mIoU} & \textbf{F1} & \textbf{Pixel Acc.} \\
\midrule
Traversability $f_{\text{trav}}$ & 0.914 & 0.964 & 0.969 \\
Frontier $f_{\text{front}}$      & 0.320 & 0.465 & 0.995 \\
\bottomrule
\end{tabular}
\end{table}

\begin{figure}[htbp]
    \centering
    \includegraphics[width=0.5\textwidth]{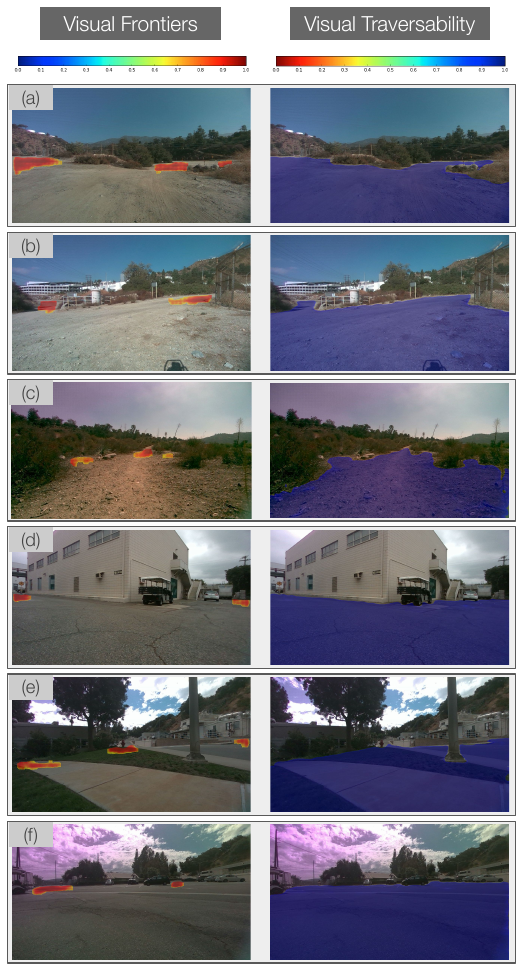}
    \caption{\textbf{Examples of Visual Traversability and Frontiers from ExploRFM in varied terrains.} Visual Frontiers are thresholded and shown in the Jet Colormap (Red indicating higher frontier confidence). Visual Traversability is also thresholded and shown in the inverse Jet Colormap (Blue indicating higher traversability confidence). (a),(b): Offoad environments with a dirt-track (c): Narrow trail amidst rocks, bushes and undergrowth (d),(e): Urban environments with paved terrains (f): Vast and open urban terrain}
    \label{fig:trav_front_examples}
\end{figure}

\paragraph{Qualitative Results.}
Figure~\ref{fig:trav_front_examples} shows qualitative outputs of ExploRFM from field deployments of WildOS across diverse environments.
The visual frontiers accurately capture semantically promising exploration cues, such as openings between trees (Fig.~\ref{fig:trav_front_examples}a), visible trail ends (Fig.~\ref{fig:trav_front_examples}b,c), and ramps or junctions (Fig.~\ref{fig:trav_front_examples}e).  
The traversability maps reliably highlight safe ground regions across both off-road and paved terrains, confirming the robustness of the trained heads and the generalizability of foundation model features. 

\subsection{Object-Search Experiments}
\label{sec:objsearch_experiments}
In this section, we address the research question:

\begin{quote}
\textit{Q1: Does the complete WildOS system enable successful end-to-end object search from language queries?}
\end{quote}

This question evaluates the system-level capability of WildOS to interpret a language-based goal, ground it visually in the environment, and autonomously navigate to the corresponding object.  

\begin{figure*}[htbp]
    \centering
    \includegraphics[width=0.8\textwidth]{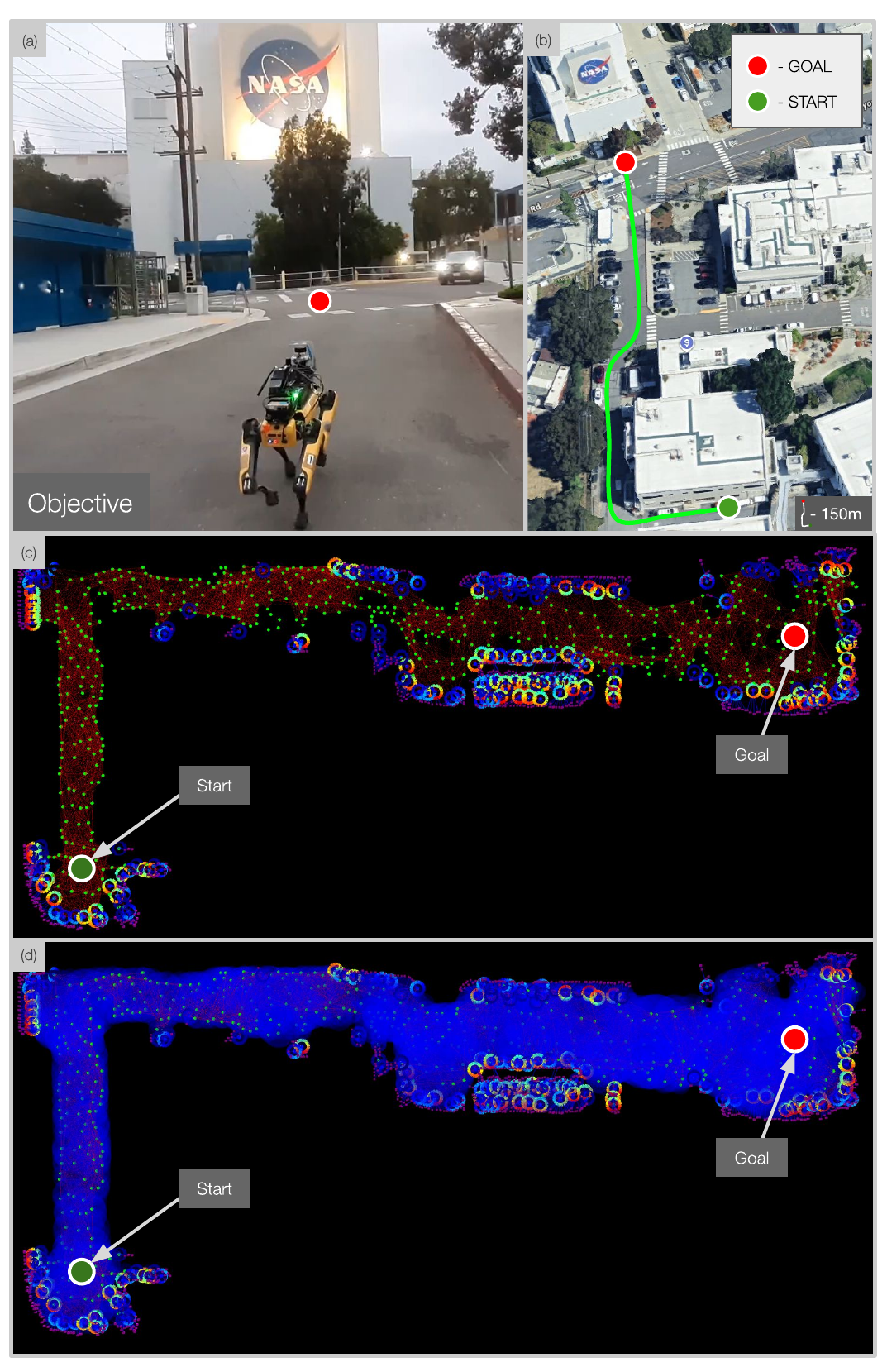}
    \caption{\textbf{Overview of the Object Search Experiment for the query - ``NASA Logo''.} (a) Third-Person view of the robot approaching the NASA sign. (b) Expected trajectory of the robot overlaid on a satellite image showing the environment. (c) Scored Navigation Graph after a successful run. The rings around the frontier nodes denote the scores for each bin according to the Jet Colormap - higher scores being red, and the lower ones being blue. (d) Explored regions of the graph - visualizing the explored radius of each node in blue.}
    \label{fig:x_logo_1}
\end{figure*}

\paragraph{Setup and Objective.}
To test open-vocabulary grounding and search capabilities, we conduct an experiment with the query \textit{``NASA logo''}, where the goal is to reach the large NASA sign located at the JPL campus.  
The robot is initialized inside an alley with an approximate prior goal $\mathbf{g}_0$ placed to the northwest of the sign.  
An overview of the mission setup is shown in Fig.~\ref{fig:x_logo_1}(a)--(b).  
The robot successfully navigates a total distance of $\sim\!150$\,m, autonomously exiting the alley, turning onto the main road, and stopping directly beneath the NASA sign.

\begin{figure*}[htbp]
    \centering
    \includegraphics[width=\textwidth]{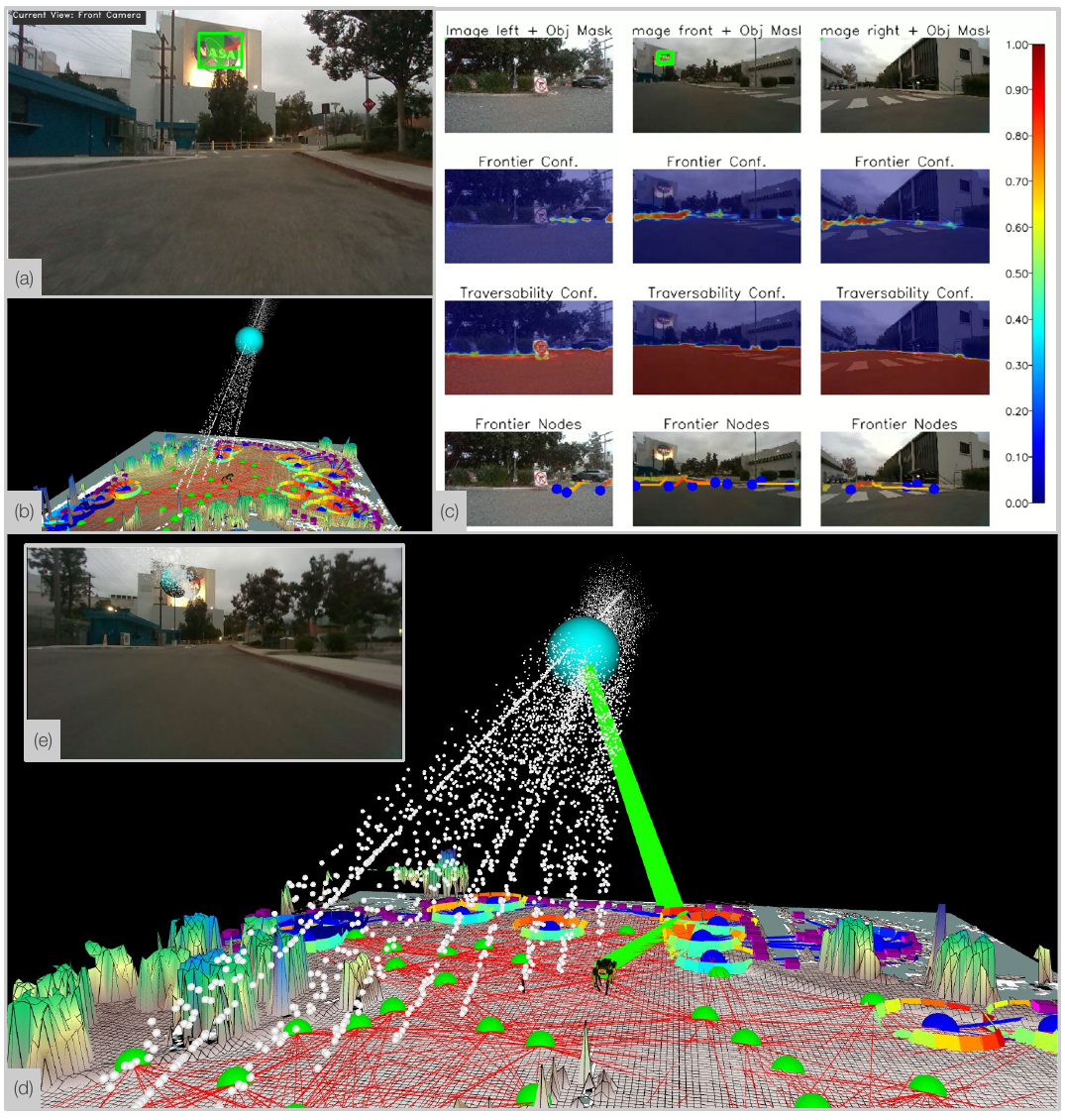}
    \caption{\textbf{ExploRFM Outputs and Triangulation of ``NASA Logo''.} (a) Object Similarity map from ExploRFM (b) Initial triangulation of the NASA sign (c) Visual Frontiers, Traversability and Projected frontier node scores for all three cameras. Frontier and Traversability Confidence is shown using the Jet Colormap. (d) Final Triangulation of the logo (e) RViZ markers projected in the front camera}
    \label{fig:x_logo_2}
\end{figure*}

During navigation, the system continuously scores frontier nodes along the edge of the local costmap using the visual traversability $\mathcal{T}^{\text{vis}}_t$ and frontier maps $\mathcal{F}^{\text{vis}}_t$ predicted by ExploRFM (Fig.~\ref{fig:x_logo_2}(c)).  
Once the target object becomes visible, it is detected through the object-similarity map $\mathcal{S}^{\text{vis}}_t$ (Fig.~\ref{fig:x_logo_2}(a)).  
Particles are sampled and projected following the coarse triangulation procedure described in Sec.~\ref{sec:triangulation}, producing an initial goal hypothesis (Fig.~\ref{fig:x_logo_2}(b)).  
The active navigation goal is then switched from $\mathbf{g}_0$ to the triangulated estimate $\hat{\mathbf{p}}_{\text{goal}}$.

\paragraph{Results and Analysis.}
A visualization of all projected particles across multiple views of the object is shown in Fig.~\ref{fig:x_logo_2}(d).  
When projecting the estimated goal marker (blue sphere) from the global map into the camera frame (Fig.~\ref{fig:x_logo_2}(e)), it aligns accurately with the NASA sign, indicating a precise triangulation.  
The complete scored navigation graph $\mathcal{G}^{\text{score}}_t$ at the end of the run is visualized in Fig.~\ref{fig:x_logo_1}(c), preserving traversability information from the local costmap $\mathcal{T}^{\text{geo}}_t$ over time.  
Fig.~\ref{fig:x_logo_1}(d) further shows the explored radius $r^e_i$ of each graph node, denoted by blue circles, effectively summarizing the explored area during the mission.

\begin{figure*}[htbp]
    \centering
    \includegraphics[width=\textwidth]{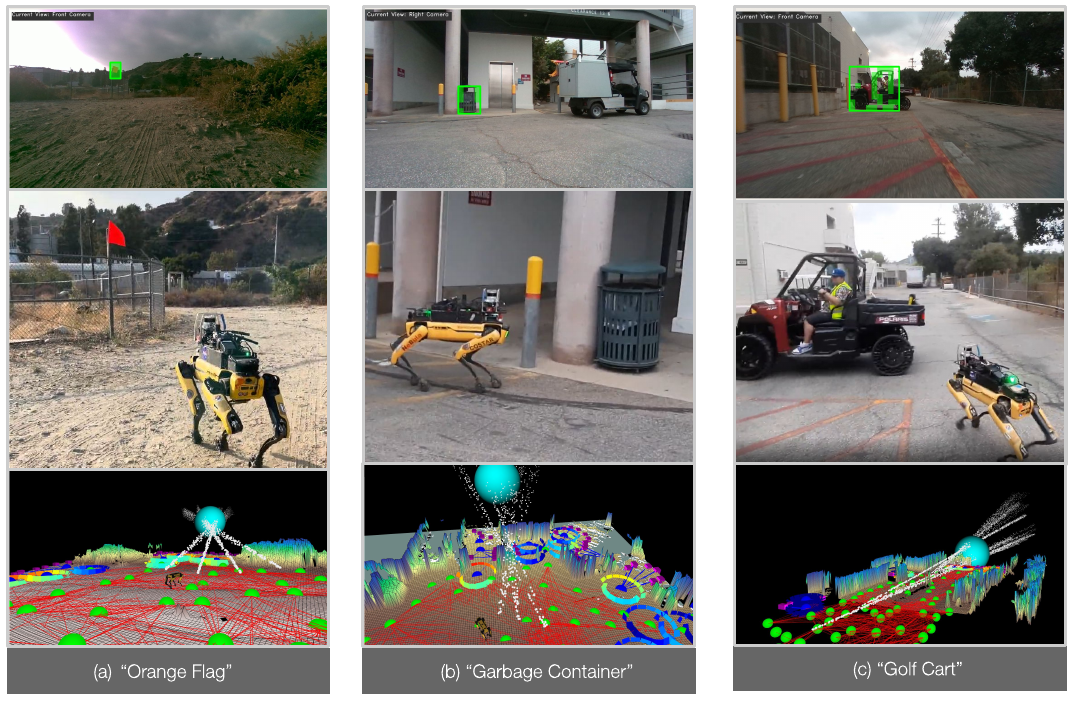}
    \caption{\textbf{More Object-Search Experiments} (a)"Orange Flag" (b)"Garbage Container" (c)"Golf Cart"}
    \label{fig:obj_search_exps}
\end{figure*}

\paragraph{Zero-Shot Generalization.}
To further evaluate the generality of WildOS, we conduct additional zero-shot search runs with the language queries:  
\textit{``orange flag''}, \textit{``garbage container''}, and \textit{``golf cart''}.  
Figure~\ref{fig:obj_search_exps} illustrates qualitative results across these categories.  
Even for cases with limited viewing baselines (e.g., the golf cart in Fig.~\ref{fig:obj_search_exps}(c)), the system produces robust coarse localization estimates, demonstrating the effectiveness of the particle-based triangulation.

\par These experiments demonstrate that the system can ground arbitrary language queries into visual similarity maps, estimate coarse 3D goal locations through triangulation, and autonomously navigate toward them. Together, they highlight the integration of language grounding, vision-based localization, and geometric planning operating in real time on a deployed platform.

\subsection{Long-Range Navigation Experiments}
\label{sec:lrn_exps}
In this section, we isolate the navigation component of WildOS to evaluate its ability to perform goal-directed traversal \emph{without} object search.  
Here, the objective is to reach a predefined point goal $\mathbf{g}_0$ provided at the start of the mission.  
Unlike the object-search experiments, we disable the object-similarity head and do not compute $\mathcal{S}^{\text{vis}}_t$ for these trials.  
These experiments allow us to analyze how the integration of vision-based reasoning and graph-based memory contributes to robust long-range navigation.

\subsubsection{Baselines}

\paragraph{Long Range Navigation (LRN).}
LRN~\cite{schmittle2025lrn} is a purely vision-based frontier navigation method.  
It discretizes the space around the robot into a fixed set of angular bins and uses the visual frontier heatmaps $\mathcal{F}^{\text{vis}}_t$ from each camera to assign scores to these bins using the known camera intrinsics.  
Scores from overlapping views are merged by taking the maximum, and bins below a confidence threshold are suppressed.  
To ensure goal-directed behavior, LRN computes two other scores: it conditions on the relative heading of the goal and smooths trajectories by also considering the robot’s previous heading.  
The resulting weighted score vectors are combined multiplicatively, and the bin with the highest final score determines the next navigation heading.  
The local planner receives a short-range goal at the edge of the costmap in this chosen direction.  
LRN operates entirely in the image space and does not maintain any spatial memory — it does not record previously explored regions or deferred frontier options.
To ensure a fair comparison at the navigation level, we replace LRN’s original frontier detector with ExploRFM and use its visual frontier predictions as input to LRN.

\paragraph{Vanilla GraphNav.}
This baseline leverages the navigation graph but removes the influence of vision-based scoring on planning.  
Specifically, we modify the scaling function in Eq.~\ref{eq:planning_scale_func} to a constant, $z(s_i) = 2$, resulting in uniform edge weighting.  
A small penalty is applied for exploring beyond the current graph, making the auxiliary goal node edges twice their Euclidean distance.  
This setup mirrors traditional geometry-based exploration and planning, relying solely on the graph’s connectivity and geometry for navigation decisions.

\subsubsection{Evaluating Efficiency of Vision-Based Scoring}
\label{sec:q2}

We now address the question:  
\begin{quote}
\textit{Q2: Does integrating vision-based scoring with the navigation graph improve navigation performance compared to pure-geometry or pure-vision approaches?}
\end{quote}

\begin{figure*}[htbp]
    \centering
    \includegraphics[width=\textwidth]{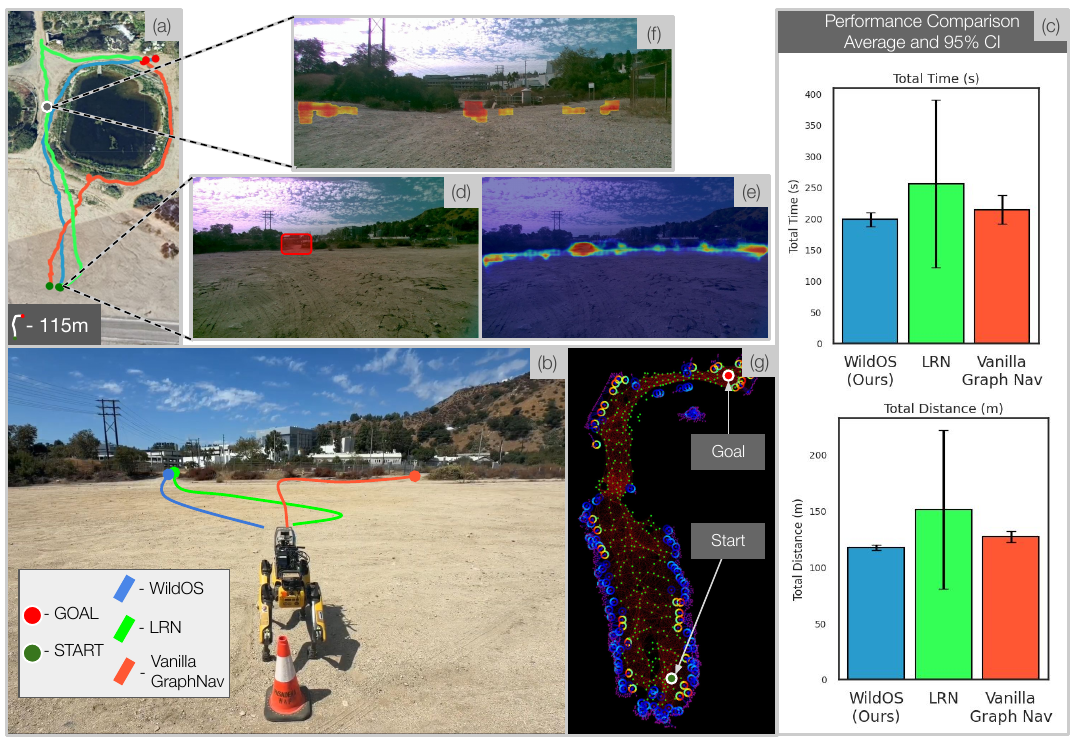}
    \caption{\textbf{Evaluating Vision-Based Scoring for Navigation Efficiency.}
    (a) GPS trajectories for all methods.  
    (b) Third-person view at the start of the experiment with future paths overlaid.  
    Vision-based approaches detect the corridor opening early on and move towards it, whereas Vanilla GraphNav heads straight towards the goal until it reaches the fence and goes around.  
    (c) Quantitative comparison across three runs (mean $\pm$ 95\% CI); the underlying values are listed in Table~\ref{tab:q2_results}.  
    (d), (e) Visual frontier and the confidence map at the start of the experiment $\mathcal{F}^{\text{vis}}_0$, highlighting the opening between the fences.  
    (f) Tri-fork after exiting the corridor, where LRN~\cite{schmittle2025lrn} hesitates between competing frontiers.
    (g) Scored Navigation Graph after a successful WildOS run.}
    \label{fig:approach_exp}
\end{figure*}

\paragraph{Experiment Setup.}
The robot begins approximately \SI{50}{m} in front of a circular fence, with the goal $\mathbf{g}_0$ located behind it as shown in Fig.~\ref{fig:approach_exp}(a),(b).  
A second circular fence lies adjacent (left) to the first, creating a narrow corridor between them.  
The surrounding area is a wide open field with no other obstacles.  
Each method is evaluated over three independent runs from the same start pose, and we report the total trajectory length, the traversal time, and the number of operator interventions--the three proxies for $J(\xi)$ introduced in Sec.~\ref{sec:problem_statement}.  
Trajectory length is integrated from the odometry estimate and time is measured from mission start until the goal acceptance radius $d_{\textit{reach}}$ is entered.

\paragraph{Qualitative Observations.}
Figure~\ref{fig:approach_exp}(a) visualizes the GPS trajectories for all methods.  
Both vision-guided approaches, LRN and WildOS, immediately identify the corridor between the fences and plan paths toward it.  
As shown in Fig.~\ref{fig:approach_exp}(e), the visual frontier map $\mathcal{F}^{\text{vis}}_0$ assigns high confidence to the pixels corresponding to this opening, indicating a strong frontier likelihood.  
This allows WildOS to prioritize the opening rather than following the straight (but blocked) direction toward the goal--effectively reasoning about accessibility rather than mere goal direction. This replicates human-like reasoning that prefers \emph{affordable} directions over direct ones.

In contrast, \textit{Vanilla GraphNav} proceeds straight toward the fence until it appears in the local map, at which point the front-facing frontier nodes are cleared, forcing it to detour around the obstacle.  
This illustrates its myopic, locally reactive behavior.  

\paragraph{Quantitative Results.}
Table~\ref{tab:q2_results} summarizes the quantitative results. WildOS achieves the lowest average distance and time across three runs and exhibits notably smaller variance compared to both baselines, shortening the traversal by $7.8\%$ in distance and $7.1\%$ in time relative to Vanilla GraphNav, and by $22.5\%$ and $22.3\%$ relative to LRN.  
No method required an operator intervention in this scenario.
This improvement is directly explained by the reachability confidence term $R_{\text{conf}}$ in Eq.~\ref{eq:scoring_eq}, which assigns higher scores to frontier nodes whose projected paths in image space are shorter to the nearest visual frontier. In this experiment, nodes pointing toward the opening between the fences received higher scores than those directly facing the fence or the goal, enabling WildOS to plan an efficient route around the obstruction rather than heading straight toward it.

\begin{table}[htbp]
\centering
\caption{\textbf{Q2 -- Navigation efficiency in the fence-corridor scenario.} Mean $\pm$ 95\% confidence interval over three runs per method; these are the values plotted in Fig.~\ref{fig:approach_exp}(c). \emph{Interv.} is the total number of operator interventions across the three runs.}
\label{tab:q2_results}
\renewcommand{\arraystretch}{1.2}
\setlength{\tabcolsep}{5pt}
\begin{tabular}{lccc}
\toprule
\textbf{Method} & \textbf{Distance ($\downarrow$) [m]} & \textbf{Time ($\downarrow$) [s]} & \textbf{Interv. ($\downarrow$)} \\
\midrule
LRN~\cite{schmittle2025lrn} & $150.7 \pm 70.3$ & $255.6 \pm 134.4$ & 0 \\
Vanilla GraphNav           & $126.6 \pm 5.0$  & $213.9 \pm 22.9$  & 0 \\
\rowcolor{green!20}
\textbf{WildOS (Ours)}     & $\mathbf{116.7 \pm 2.3}$ & $\mathbf{198.6 \pm 10.9}$ & 0 \\
\bottomrule
\end{tabular}
\end{table}

\paragraph{Analysis of LRN Behavior.}
Interestingly, LRN performs worse than even the geometry-only baseline.  
As shown in the trajectories of Fig.~\ref{fig:approach_exp}(a), LRN initially selects the right-hand opening between the fences but later switches to the left, demonstrating oscillatory behavior due to varying frontier activation values - a limitation also noted by its authors~\cite{schmittle2025lrn}.  
Further, near the exit of the corridor, LRN encounters a tri-fork with three visible frontiers (Fig.~\ref{fig:approach_exp}(f)).  
Although the rightmost frontier leads directly to the goal, in two out of three runs LRN continues forward before eventually turning right.  
This can be explained by referencing the method LRN uses to aggregate frontier activations: it sums all pixel scores within an angular bin and normalizes them, favoring frontiers that are larger in image space.  
In Fig.~\ref{fig:approach_exp}(f), the frontier leading toward the goal appears smaller due to perspective, thus receiving a lower aggregate score and causing suboptimal navigation choices.

\begin{figure}[htbp]
    \centering
    \includegraphics[width=0.5\textwidth]{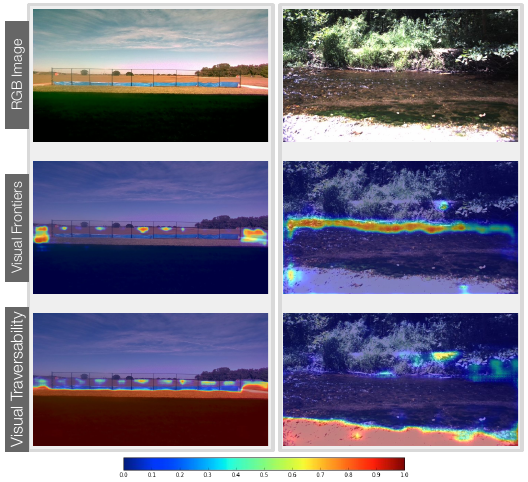}
    \caption{\textbf{Potential LRN Failure Cases.} We visualize the visual frontier and traversability heatmaps (unthresholded) overlaid on top of the RGB image - both in the Jet Colormap i.e. red indicating a higher probability score. Notice how relying purely on frontier heatmaps (LRN~\cite{schmittle2025lrn}) can lead to high affordance scores for non-reachable frontiers. WildOS tackles this by incorporating visual traversability in frontier scoring.}
    \label{fig:lrn-trav-fail}
\end{figure}

Another shortcoming of LRN~\cite{schmittle2025lrn} is that it assumes implicit reachability for the visual frontiers and the frontier detection module does not explicitly account for traversability information. If there were visual frontiers detected beyond non-traversable regions the robot would still score the angular bins in those directions which is inefficient. It would plan around only when the non-traversable region appears within its local cost map. Fig~\ref{fig:lrn-trav-fail} showcases potential examples with the visual frontier heatmaps and the traversability heatmaps visualized. In contrast to LRN, WildOS explicitly incorporates visual traversability into frontier scoring, allowing it to suppress visually salient but unreachable regions and avoid pursuing such \emph{non-reachable} frontiers.

\medskip

\subsubsection{Evaluating Efficiency of a Navigation Graph}
\label{sec:q3_deadend}

\begin{quote}
\textit{Q3: Does the navigation graph improve robustness and memory compared to purely vision-based navigation?}
\end{quote}

\paragraph{Experiment Setup.}
This experiment isolates the contribution of the navigation graph in handling ambiguous or partially explored environments. We design a scenario involving a dead-end to evaluate the ability of each method to remember explored regions and recover from failure.

\begin{figure*}[htbp]
    \centering
    \includegraphics[width=\textwidth]{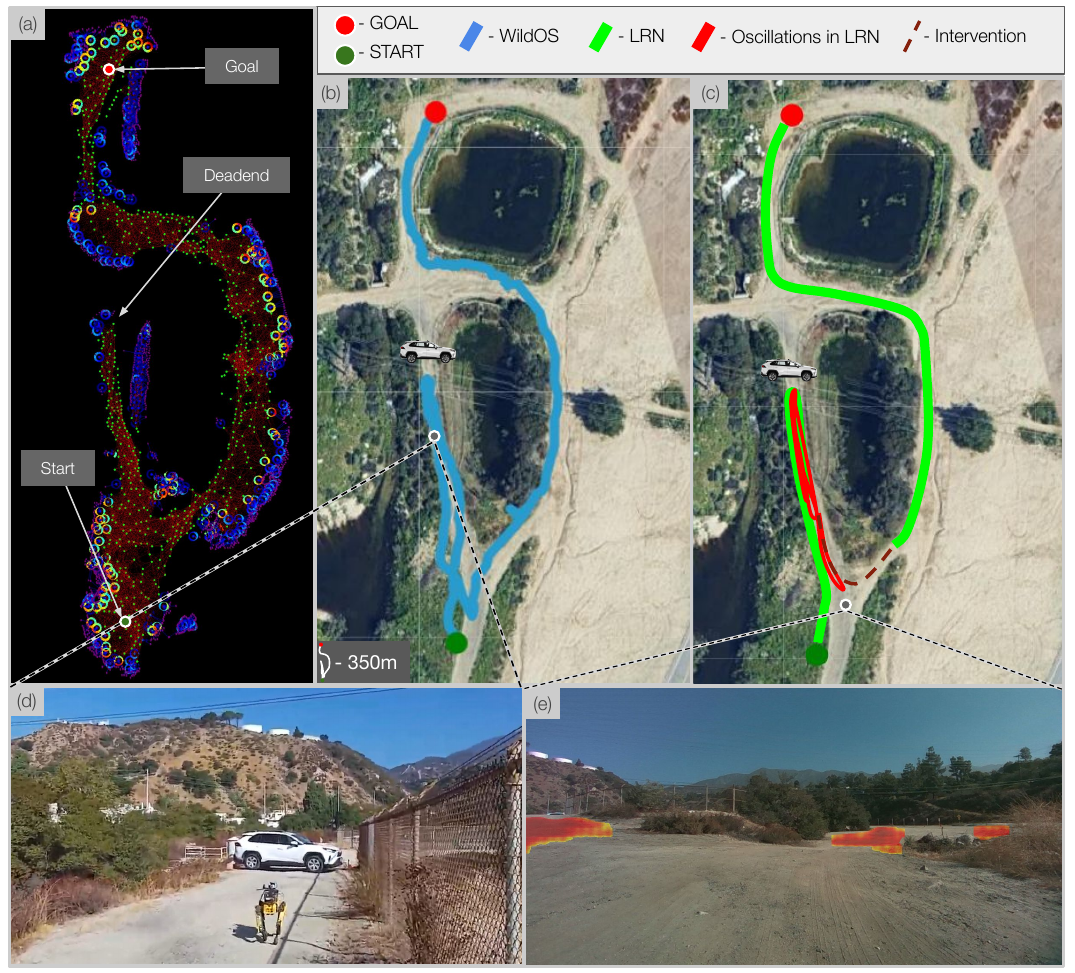}
    \caption{\textbf{Evaluating the Role of the Navigation Graph in Handling dead-ends.}
    (a) Final scored navigation graph visualized after a successful WildOS run.
    (b) GPS trajectory for WildOS showing successful rerouting around the fence.  
    (c) GPS trajectory for LRN, where human intervention was required. Quantitative results for both methods are listed in Table~\ref{tab:q3_results}.  
    (d) Third-person view showing the blocked trail.
    (e) Visual frontiers at the fork - WildOS initially takes the opening on the left given its better alignment with the goal. After encountering a dead-end it turns back and takes the opening on the right. LRN does not store that it has already explored the opening on the left and keeps choosing the opening on the left at this fork and turning back due to a dead-end leading to an oscillatory behavior.
    }
    \label{fig:deadend_exp}
\end{figure*}

\medskip
The robot is initialized in front of an elliptical fence with a fork ahead, offering two possible trails (Fig.~\ref{fig:deadend_exp}(b),(c)). The left trail appears shorter and more aligned with the goal direction; however, further down, we block this path using a parked car (Fig.~\ref{fig:deadend_exp}(d)), creating a dead-end. Ideally, a robust navigation system should recognize the blockage, turn back, and select the alternate, longer path around the fence to reach the goal.

We compare WildOS and LRN over three runs each. Vanilla GraphNav is not included here: it shares the navigation graph with WildOS and therefore recovers from the dead-end through the same mechanism, whereas the scenario is constructed to isolate persistent memory, the axis along which LRN differs.

\paragraph{Intervention Protocol and Failure Criterion.}
An \emph{intervention} is declared when the robot stops reducing its distance to the goal for more than \SI{60}{s} while remaining within a \SI{10}{m} radius, that is, when it has settled into a persistent oscillation rather than a deliberate detour.  
The safety operator then takes manual control, repositions the robot roughly \SI{5}{m} along the branch the system failed to commit to, and hands autonomy back.  
The reported distance and time include the assisted segment and the autonomous recovery that follows it.  
A run is declared \emph{failed} if the goal is not reached within \SI{1200}{s} of mission time, or if the operator has to stop the robot to prevent a collision; no run in this experiment terminated as a failure under this criterion.

\paragraph{Qualitative Observations.} WildOS consistently completes the mission successfully in all runs, while LRN requires human intervention every time. WildOS initially chooses the shorter path, reaches the dead-end, and upon clearing the local frontier nodes, reroutes using the navigation graph. Frontier nodes pointing toward the unexplored longer path were assigned higher scores since they were visually rich directions to explore (Fig.~\ref{fig:deadend_exp}(e)), enabling the system to autonomously turn back and reroute around the fence to reach the goal. The complete scored navigation graph after the run is visualized in Fig.~\ref{fig:deadend_exp}(a).

\medskip
In contrast, LRN, being purely vision-based and memoryless, fails to recover. Upon reaching the dead-end, it observes no visible frontiers and begins oscillating left and right. Eventually it turns back and attempts to take the longer path. However, it redetects the same visual frontier on the blocked path in its left camera and repeatedly chooses it, leading to a cyclic behavior. Only after manual intervention--where the robot is repositioned partway down the longer path--does LRN successfully complete the run (Fig.~\ref{fig:deadend_exp}(c)).

\paragraph{Quantitative Results.}
Table~\ref{tab:q3_results} summarizes the three runs per method.  
WildOS completes the mission autonomously in all three runs, while LRN triggers the intervention criterion exactly once per run.  
Without the intervention it does not terminate at all, so the autonomy gap rather than the efficiency gap is the meaningful result of this experiment.

\begin{table}[htbp]
\centering
\caption{\textbf{Q3 -- Dead-end recovery.} Mean $\pm$ 95\% confidence interval over three runs per method. \emph{Interv.} is the total number of operator interventions across the three runs; distance and time include the assisted segment and the autonomous recovery that follows it.}
\label{tab:q3_results}
\renewcommand{\arraystretch}{1.2}
\setlength{\tabcolsep}{5pt}
\begin{tabular}{lccc}
\toprule
\textbf{Method} & \textbf{Distance ($\downarrow$) [m]} & \textbf{Time ($\downarrow$) [s]} & \textbf{Interv. ($\downarrow$)} \\
\midrule
LRN~\cite{schmittle2025lrn} & $439.0 \pm 89.8$ & $791.3 \pm 71.4$ & 3 \\
\rowcolor{green!20}
\textbf{WildOS (Ours)}      & $\mathbf{345.9 \pm 44.1}$ & $\mathbf{636.7 \pm 120.4}$ & \textbf{0} \\
\bottomrule
\end{tabular}
\end{table}

\medskip
These results demonstrate that persistent spatial memory is essential for long-horizon autonomy. By maintaining a structured representation of previously explored regions and deferred frontiers, WildOS can recover from dead-ends and replan effectively, whereas memoryless vision-only strategies remain prone to oscillation and repeated failure.

\subsubsection{Generalization across terrains}
\label{sec:q4_generalization}
\begin{quote}
\textit{Q4: Does WildOS generalize effectively across diverse outdoor terrains?}
\end{quote}

\paragraph{Experiment Setup. }
This experiment evaluates the ability of WildOS to generalize to outdoor environments. Importantly, the frontier head of ExploRFM is trained solely on 350 annotated images from the GrandTour~\cite{grandtour} dataset, which spans varied environments. We perform no additional fine-tuning or adaptation on data from the deployment environments.

While all prior experiments were conducted in off-road terrains, here we evaluate navigation performance in urban environments.

\begin{figure*}[htbp]
    \centering
    \includegraphics[width=\textwidth]{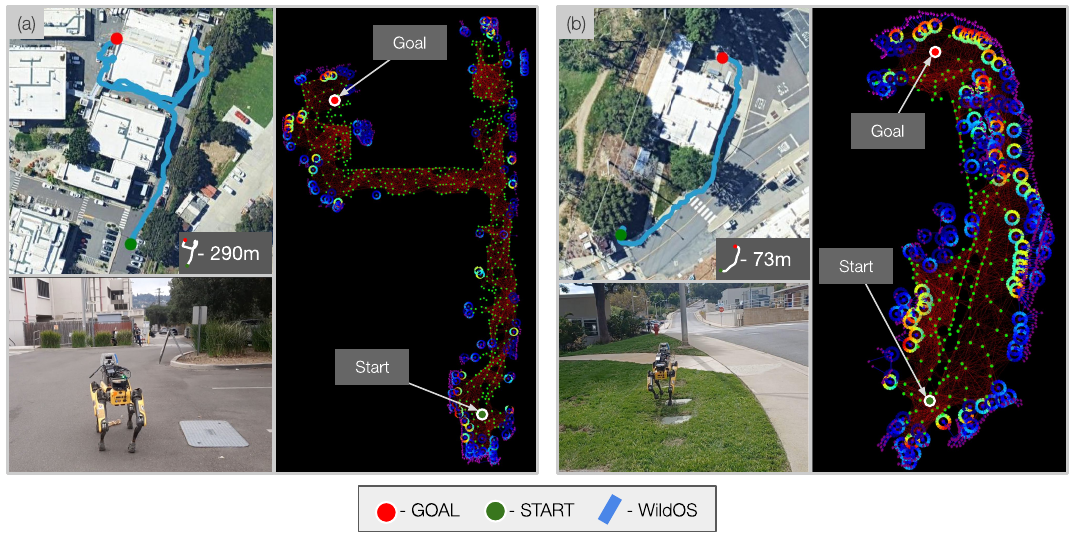}
    \caption{\textbf{WildOS in Urban Environments.} Third-person views of the robot at the start of the run, GPS trajectories, and scored navigation graphs at the end of the runs for two scenarios:  
    (a) Goal located within a cluster of buildings.  
    (b) Goal located behind a building.  
    WildOS plans efficient routes around obstacles and navigates to the goal successfully.}
    \label{fig:urban_exp}
\end{figure*}

\paragraph{Qualitative Observations. }
In the first experiment (Fig.~\ref{fig:urban_exp}(a)), the goal is placed in the middle of a cluster of buildings. The robot initially proceeds along the straight road toward the northernmost building but detects a blockage. It subsequently uses the navigation graph to turn around, and reroute through an alleyway between two buildings to reach the goal successfully.

\medskip
In the second experiment (Fig.~\ref{fig:urban_exp}(b)), the robot starts in front of a building with the goal located directly behind it. Although a direct path leads inside the building, WildOS correctly provides low scores to frontier nodes pointing to the direct path, instead selecting a path that cleanly circumvents the building. 

\medskip
These experiments highlight the strong generalization of WildOS, enabled by foundation-model features - allowing it to adapt seamlessly across terrain types without any retraining or environment-specific tuning.

\subsection{Discussion}

\begin{table}[H]
    \centering
    \caption{\textbf{Comparison of Navigation Capabilities.}}
    \label{tab:baseline_capabilities}
    \renewcommand{\arraystretch}{1.2}
    \setlength{\tabcolsep}{6pt}
    \resizebox{\columnwidth}{!}{%
    \begin{tabular}{lcc}
        \toprule
        \textbf{Method} & \textbf{Vision-based Reasoning} & \textbf{Graph Memory / History} \\
        \midrule
        LRN~\cite{schmittle2025lrn} & \checkmark & \xmark \\
        Vanilla GraphNav & \xmark & \checkmark \\
        \rowcolor{green!20}
        \textbf{WildOS (Ours)} & \checkmark & \checkmark \\
        \bottomrule
    \end{tabular}%
    }
\end{table}

The three approaches in Table~\ref{tab:baseline_capabilities} collectively span the design space of navigation paradigms:  
\emph{pure vision} (LRN), \emph{pure geometry} (Vanilla GraphNav), and our proposed \emph{vision + memory} (WildOS). The results we showcase in this work consistently show that combining vision-based reasoning with geometric memory yields the most robust and efficient navigation behavior.

\subsubsection{Robustness to False Positives and Imperfect Object Detections}
\label{sec:detection_robustness}

WildOS mitigates false positives through accumulation of evidence across viewpoints. The open-vocabulary similarity map $\mathcal{S}^{\text{vis}}_t$ is consumed only by the triangulation module, where each observation contributes a bounding-box ray and particles inconsistent with multiple rays receive vanishing weight. As the robot approaches the target, the increasing viewpoint baseline further reduces localization error and corrects residual bias in $\hat{\mathbf p}_{\text{goal}}$. 
As quantified in Appendix~\ref{app:triangulation}, the goal estimate tolerates bounding-box noise up to \SI{16}{px} and its error decays as $1/\sqrt{N_c}$ with accumulated views. The primary failure mode is therefore persistent model bias across viewpoints rather than transient hallucinations. Even then, the impact is strictly bounded: semantic scores only reweight path costs in Eq.~\eqref{eq:planning_cost} and cannot make non-traversable edges traversable. Incorrect predictions degrade only exploration efficiency, while safety remains guaranteed by the geometric navigation graph and local elevation map.

Frontier scoring is robust to isolated semantic errors since the final score is the \emph{product} of three confidence terms $G_{\text{conf}} \cdot R_{\text{conf}} \cdot F_{\text{conf}}$, each bounded in $[0,1]$: preventing a single high-confidence false positive from dominating the decision process. The primary remaining failure mode is persistent bias in the underlying foundation model, where the same object is misclassified across many viewpoints. With modern open-vocabulary VLMs this regime has become noticeably less frequent. Even in this case, geometric safety is preserved because navigation remains constrained to the navigation graph and local traversability map. Conversely, if the target is missed entirely due to distance, occlusion, or ambiguity, the robot continues navigating toward the coarse prior goal $\mathbf{g}_0$ until sufficient observations are available for re-detection and refinement.
\section{CONCLUSION, LIMITATIONS \& FUTURE WORK}
\subsection{Conclusion}

In this work, we presented WildOS, a unified, real-time system for open-vocabulary object search and long-range navigation in unstructured outdoor environments. Our system combines a navigation graph with foundation-model-based perception to achieve semantic, memory-aware exploration at scale. By integrating the geometric memory of a navigation graph with the semantic and visual priors from ExploRFM, WildOS is able to plan globally while making visually meaningful local decisions.

We showcased the strength of foundation-model features in robotics, demonstrating generalization across terrains and robust performance with minimal training data (only 350 annotated images). Through extensive real-world experiments, WildOS outperformed SOTA long range navigation baselines in complex scenarios such as dead-ends and occluded goal regions.  
This work highlights a step toward bridging vision-language reasoning and long-horizon autonomy--paving the way for more general, goal-driven robotic exploration in the wild.

\subsection{Limitations}

While WildOS demonstrates strong performance, several limitations remain that open directions for further improvement:

\begin{itemize}
    \item \textbf{Oscillatory behavior in long-horizon exploration.}  
    In environments with multiple high-scoring but spatially distant frontier nodes, the planner can occasionally oscillate between far-apart regions when frontier scores fluctuate slightly. This occurs in large, complex areas such as long trails with multiple narrow exits, some of which may be blocked. A more principled mechanism for balancing exploration and exploitation--allowing the system to commit to a chosen region until it is fully explored--could mitigate such inefficiencies.
    
    \item \textbf{Lack of visual feature memory for object search.}  
    Currently, WildOS does not store visual features (e.g., embeddings from ExploRFM) at graph nodes. This limits the system’s ability to perform retrospective object search—e.g., identifying that a previously seen object matches a newly issued language query.
    
    \item \textbf{Approximate explored-region geometry.}  
    The current implementation assumes that each explored region around a node is circular, which simplifies storage but can leave small unmarked regions between overlapping circles. In cluttered environments (e.g., dense vegetation or uneven terrain), this can lead to spurious frontier detections near already explored areas eventually making the graph unstable.
    
    \item \textbf{Uniform threshold for open-vocabulary segmentation.}  
    We use a single similarity threshold across all object categories during open-vocabulary grounding. However, we observe that optimal thresholds vary between categories. A category-adaptive thresholding mechanism could enhance detection precision and reduce false positives during language-conditioned search.

    \item \textbf{Coarse goal prior at long range.}  
    The triangulated estimate $\hat{\mathbf{p}}_{\text{goal}}$ carries a position error that grows roughly linearly with target distance and is concentrated along the line of sight (Appendix~\ref{app:triangulation}). This is sufficient for biasing exploration toward the target, but the estimate cannot serve as a terminal goal until the object enters range sensing.

    \item \textbf{Empirically selected scoring parameters.}  
    The functional forms of the confidence terms in Eq.~\eqref{eq:scoring_eq}, the scaling function $z(\cdot)$ of Eq.~\eqref{eq:planning_scale_func}, and the thresholds of Table~\ref{tab:system_params} were chosen empirically during field deployment and held fixed thereafter. The bounded, multiplicative structure of the score makes the system tolerant to moderate changes in any single term, but we have not characterized closed-loop sensitivity to these choices.

    \item \textbf{Scope of field evaluation.} 
    Conducting repeated trials in large-scale outdoor environments introduces practical constraints on trial counts; in our case, the subsequent unavailability of the experimental platform bounded our evaluation to the reported field campaign. Consequently, our quantitative comparisons are evaluated over three runs per method across representative scenarios, where the reported confidence intervals characterize empirical run-to-run variation rather than formal statistical bounds. While our baseline comparisons systematically isolate the core axes of vision guidance and graph memory (Table~\ref{tab:baseline_capabilities}), exhaustive on-hardware ablations remain an area for future work. Additionally, our field trials focused exclusively on ground-level outdoor terrain, leaving multi-layer environments with vertically stacked structures, such as bridges and overpasses, untested.
\end{itemize}

\subsection{Future Work}

Building upon these findings, we outline several promising directions to extend WildOS:

\begin{itemize}
    \item \textbf{Incorporating visual memory for query-driven retrieval.}  
    One may associate \textit{CLS} embeddings from all camera views (front, left, right) with corresponding graph nodes, enabling query-based retrieval of past visual observations. By reusing these stored embeddings for triangulation--similar in spirit to TagMap~\cite{zhang2024tagmap}--the system could perform efficient open-vocabulary object mapping in large outdoor environments without requiring depth sensors.
    
    \item \textbf{Learning frontier quality for adaptive exploration.}  
    Frontier labels could be augmented with a quality or uncertainty score to allow the system to distinguish between reliable and risky exploration opportunities. This would enable more informed decision-making in environments with ambiguous paths.
    
    \item \textbf{Integrating visual and geometric traversability.}  
    Currently, visual traversability is used only beyond the local costmap. A tighter integration--projecting visual traversability into the geometric traversability map--could yield a unified local representation that leverages both semantic and geometric cues. Such a hybrid map would improve planning in environments where LiDAR-based geometry alone might be insufficient like water hazards.
\end{itemize}

\bibliographystyle{IEEEtran}
\bibliography{bibliography}

\vfill\pagebreak

\appendices

\section{Triangulation Sensitivity Analysis}
\label{app:triangulation}

The goal triangulation module of Sec.~\ref{sec:triangulation} is an auxiliary component of WildOS: it supplies a coarse, semantically driven prior on $\hat{\mathbf{p}}_{\text{goal}}$ rather than a precise localizer.  
To characterize the regime in which such a prior is useful, and in particular how its uncertainty scales with target distance, viewpoint baseline, mask noise, and particle budget, we evaluate a standalone reimplementation of the same estimator in a controlled simulation.

\subsection*{Setup}
A cubic target of side \SI{10}{m} is placed at the origin and observed by $N_c$ pinhole cameras ($640\!\times\!480$, $\mathrm{FoV}\!\approx\!90^\circ$) placed at random within a configurable distance range and azimuthal spread around the target.  
Each camera observes a noisy bounding box obtained by projecting the cube and corrupting its corners with Gaussian pixel noise ($\sigma_{\text{bbox}}$).  
The camera poses used for projection are perturbed with translation noise ($\sigma_t = \SI{1}{m}$) and rotational noise ($\sigma_R = 2^\circ$) to emulate the odometry drift typical of LiDAR--inertial estimators on long outdoor traverses.

For every camera, $N_p$ particles are sampled inside the bounding box and along the camera ray with uniform depth $d \sim \mathcal{U}(d_{\min}, d_{\max})$ and transformed to the world frame, following Algorithm~\ref{alg:project_particles}.  
Each particle is weighted as $w_k = \sum_i 1/(d_{ik} + \varepsilon)$, where $d_{ik}$ is the perpendicular distance from particle $k$ to the bounding-box ray of view $i$, giving the estimate $\hat{\mathbf p}_{\text{goal}}$ of Algorithm~\ref{alg:ray_triang}.

We report the position error $\|\hat{\mathbf p}_{\text{goal}} - \mathbf p_{\text{gt}}\|$ averaged over $200$ Monte-Carlo trials, along with its standard deviation $\sigma_e$ across those trials.  
The default configuration mirrors the deployment settings of Table~\ref{tab:system_params}: $N_c = 10$, $N_p = 1000$, $\sigma_{\text{bbox}} = \SI{4}{px}$, and an azimuthal spread of $30^\circ$, i.e. the cameras are sampled uniformly across the $30^\circ$ arc that the robot is assumed to span as it approaches the goal.  
In the distance sweep, $d_{\max}$ is scaled to $1.5\times$ the target distance.

\begin{table}[htbp]
\centering
\caption{\textbf{Triangulation sensitivity analysis.} Mean position error $\bar{e}$ and error standard deviation $\sigma_e$ over $200$ Monte-Carlo trials. The default configuration mirrors the deployment settings of Table~\ref{tab:system_params}; each block sweeps one parameter at a time.}
\label{tab:triangulation_appendix}
\small
\renewcommand{\arraystretch}{1.0}
\setlength{\tabcolsep}{5pt}
\begin{tabular}{l c r r}
\toprule
\textbf{Setting} & \textbf{Value} & $\bar{e}$ [m]$\downarrow$ & $\sigma_e$ [m] \\
\midrule
\multicolumn{4}{c}{\textbf{(a) Target distance} (cameras at $d_{\text{tgt}}\pm 5\%$, $d_{\max}=1.5\,d_{\text{tgt}}$)} \\
\midrule
$d_{\text{tgt}}$ & $\SI{15}{m}$  &  6.27 & 0.56 \\
$d_{\text{tgt}}$ & $\SI{25}{m}$  &  9.64 & 0.68 \\
$d_{\text{tgt}}$ & $\SI{50}{m}$  & 18.99 & 1.07 \\
$d_{\text{tgt}}$ & $\SI{75}{m}$  & 30.28 & 1.63 \\
$d_{\text{tgt}}$ & $\SI{100}{m}$ & 43.40 & 2.27 \\
$d_{\text{tgt}}$ & $\SI{150}{m}$ & 73.65 & 3.68 \\
\midrule
\multicolumn{4}{c}{\textbf{(b) Number of views} ($d_{\text{tgt}}\!\in\![60,100]\,\text{m}$)} \\
\midrule
$N_c$ & 2  & 35.12 & 8.89 \\
$N_c$ & 3  & 29.40 & 7.58 \\
$N_c$ & 5  & 21.84 & 6.13 \\
$N_c$ & 8  & 15.48 & 4.12 \\
$N_c$ & 10 & 12.74 & 3.83 \\
$N_c$ & 15 &  8.52 & 2.95 \\
\midrule
\multicolumn{4}{c}{\textbf{(c) Bounding-box pixel noise}} \\
\midrule
$\sigma_{\text{bbox}}$ & $\SI{0}{px}$  & 12.29 & 3.83 \\
$\sigma_{\text{bbox}}$ & $\SI{4}{px}$  & 12.74 & 3.83 \\
$\sigma_{\text{bbox}}$ & $\SI{8}{px}$  & 13.70 & 3.97 \\
$\sigma_{\text{bbox}}$ & $\SI{16}{px}$ & 16.40 & 5.33 \\
$\sigma_{\text{bbox}}$ & $\SI{32}{px}$ & 21.53 & 6.23 \\
\midrule
\multicolumn{4}{c}{\textbf{(d) Particle budget per view}} \\
\midrule
$N_p$ &   100 & 12.52 & 5.25 \\
$N_p$ &   500 & 12.90 & 3.93 \\
$N_p$ &  1000 & 12.74 & 3.83 \\
$N_p$ &  5000 & 12.65 & 3.97 \\
\bottomrule
\end{tabular}
\end{table}

\subsection*{Results}
Table~\ref{tab:triangulation_appendix} exposes three regularities that follow directly from the geometry of the estimator.

\noindent\textbf{(a) Distance.}  
Both the mean error and its standard deviation grow roughly linearly with target distance: $\bar{e}/d_{\text{tgt}}$ ranges from $\sim\!42\%$ at \SI{15}{m} to $\sim\!49\%$ at \SI{150}{m} for the nominal $30^\circ$ spread.  
The residual is dominated by depth ambiguity rather than by bearing error: over a $30^\circ$ arc the viewing rays intersect at a shallow angle, so the particle cloud is poorly constrained along the line of sight and well constrained across it.  
This is why $\hat{\mathbf{p}}_{\text{goal}}$ is used as a directional attractor in the planning cost of Eq.~\eqref{eq:planning_cost} and never as a terminal goal.

\noindent\textbf{(b) Viewpoint baseline.}  
Increasing $N_c$ from $2$ to $15$ reduces the error by roughly $4\times$, consistent with the $\propto 1/\sqrt{N_c}$ scaling expected of least-squares triangulation.  
In deployment the effective baseline is not fixed but grows with the traversed distance: every additional detection lowers both the mean error and its spread, so the prior becomes more informative exactly as the robot commits to it.

\noindent\textbf{(c) Particle budget and detector noise.}  
The error is essentially insensitive to $N_p$ above a few hundred particles per view, confirming that the residual uncertainty is set by viewpoint geometry rather than by Monte-Carlo resolution, and that $N_p$ need not be scaled with the size of the environment.  
Bounding-box noise degrades the estimate only mildly up to \SI{16}{px} and remains tractable at \SI{32}{px}.

Taken together, these results delimit what the module is for.  
The prior is accurate enough in \emph{heading} to bias frontier scoring toward the target from several hundred metres away, and its depth error is irrelevant once the target enters range sensing, at which point the median-depth estimate of Sec.~\ref{sec:triangulation} replaces it.  
Local safety is unaffected throughout, since it is anchored in the geometric navigation graph rather than in $\hat{\mathbf{p}}_{\text{goal}}$.

\vfill\pagebreak

\begin{IEEEbiography}[{\includegraphics[width=1in,height=1.25in,clip]{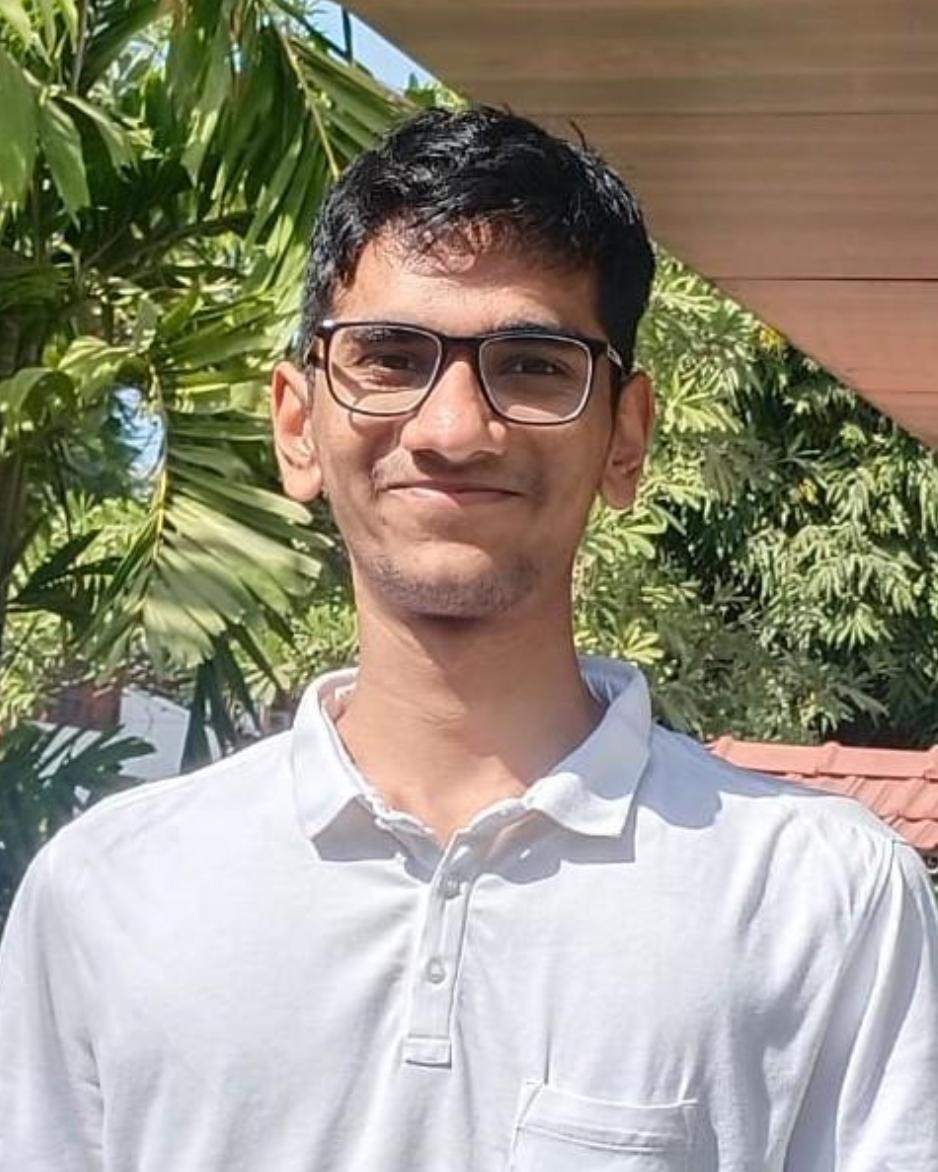}}]
{Hardik Shah}~is a Researcher at Flexion Robotics, where he works on perception for autonomous humanoid robots. He received his M.Sc. in Computer Science from ETH Zurich, Switzerland, in 2025, and B.E. degree in Computer Science from BITS Pilani, Goa, India, in 2023. His research interests include vision-based navigation and semantic mapping. He has previously conducted research at NASA's Jet Propulsion Laboratory (JPL), ETH Zurich, the University of Zurich, and Google.
\end{IEEEbiography}

\begin{IEEEbiography}[{\includegraphics[width=1in,height=1.25in,clip,keepaspectratio]{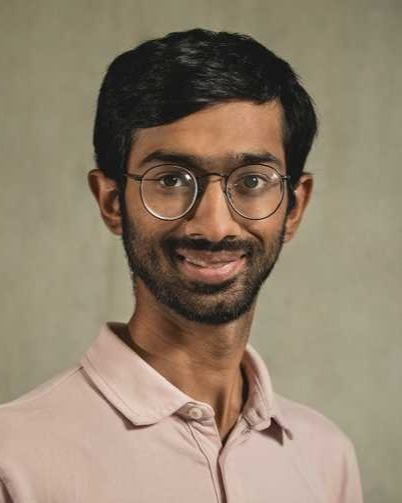}}]
{Manthan Patel}~(Student Member, IEEE)~is a PhD candidate at the Robotic Systems Lab at ETH Zurich working on foundation models for robot perception and navigation. He received his M.Sc. in Robotics, Systems \& Control in 2024 from ETH Zurich, prior to which he received a BS in Mechanical Engineering from IIT Kharagpur. His research interests lie in various aspects of field robotics with a particular focus on large-scale model training and datasets for perception and navigation.
\end{IEEEbiography} 

\begin{IEEEbiography}[{\includegraphics[width=1in,height=1.25in,clip,keepaspectratio]{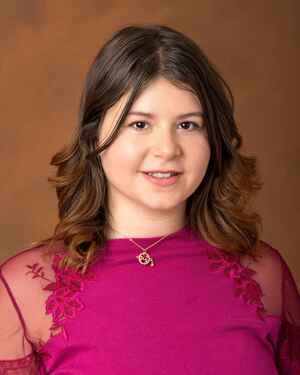}}]
{Erica Tevere}~is a Robotics Technologist in the Robotic Mobility Group at NASA’s Jet Propulsion Laboratory (JPL). She
received her Master’s in Robotics from Johns Hopkins University and her Bachelor’s of Mechanical Engineering from the University of Michigan. She currently serves as the Principal Investigator (PI) for the Hazardous Environment Robotic Mobility and Exploration Systems (HERMES) task and Autonomous Teams for Lunar Assembly and Structures (ATLAS) task and has supported projects for NASA, DARPA, USFS, and other organizations on developing and deploying autonomous solutions to fielded robotic systems.
\end{IEEEbiography} 

\begin{IEEEbiography}[{\includegraphics[width=1in,height=1.25in,clip,keepaspectratio]{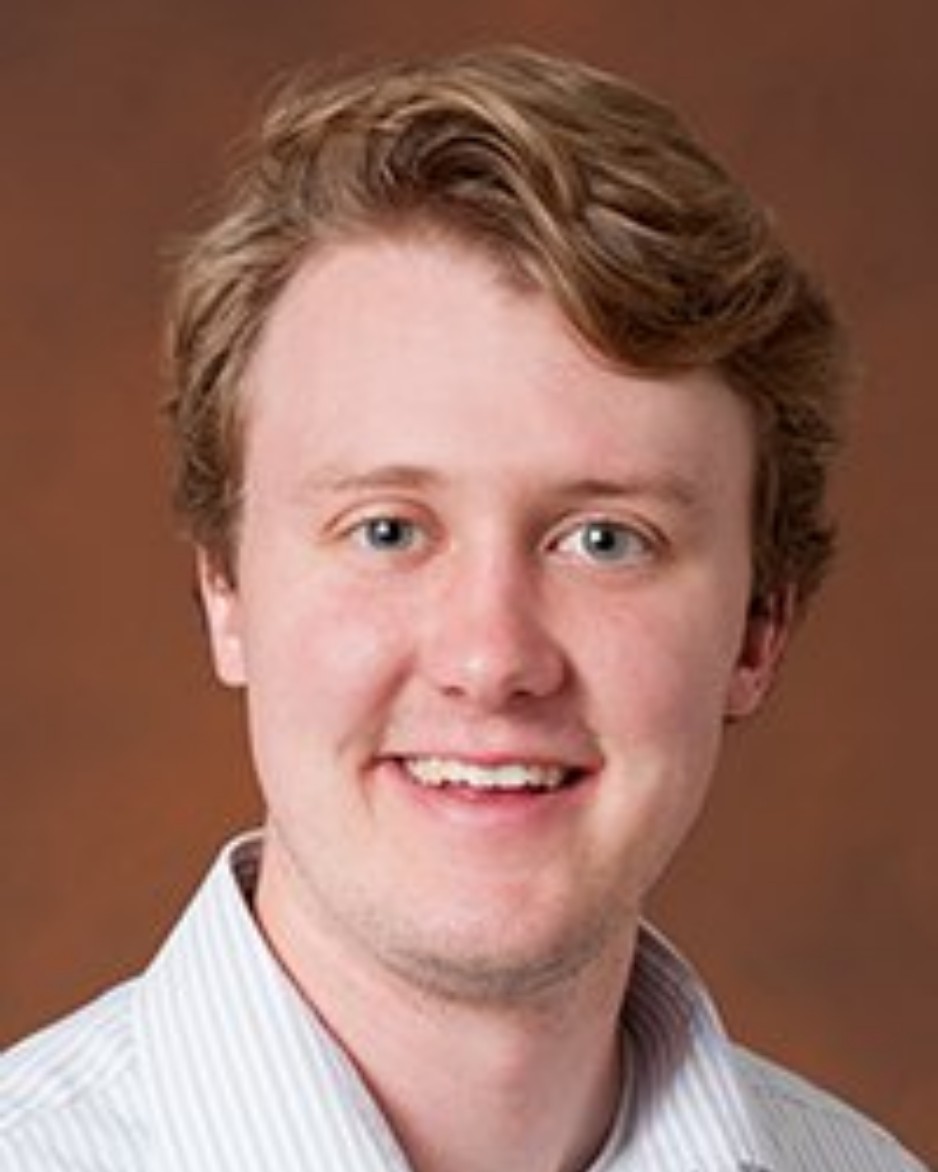}}]
{Deegan Atha}~is a Senior Autonomy Engineer in the Wheeled Autonomy Group at FieldAI. Prior to joining FieldAI, he was a Robotics Technologist in the Perception Systems Group at NASA’s Jet Propulsion Laboratory (JPL). During his time at JPL, he served as Perception Lead for JPL’s team in the DARPA RACER project and as Principal Investigator for the ShadowNav task, developing absolute global localization methods for long-distance Lunar autonomy. He holds a B.S. in Electrical Engineering from Purdue University and an M.S. in Computer Science from Georgia Tech.
\end{IEEEbiography} 

\begin{IEEEbiography}[{\includegraphics[height=1.25in,width=1in,trim=0 0 0 400pt,clip,keepaspectratio]{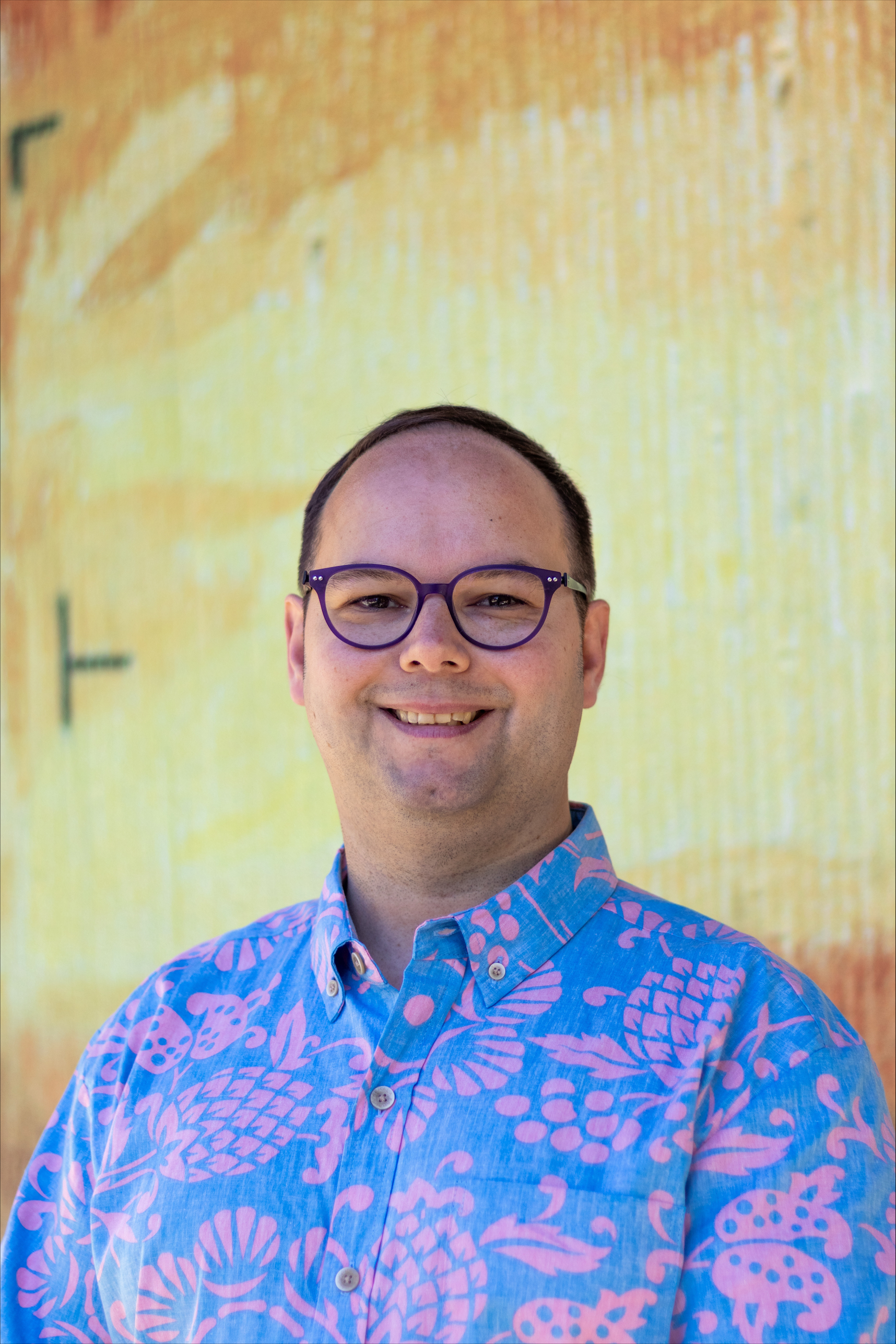}}]
{Marcel Kaufmann}~is a Robotics Researcher and Data Scientist in the Aerial Robotic Systems Group at NASA's Jet Propulsion Laboratory (JPL). His research focuses on robotics, artificial intelligence, human–robot interaction, and autonomous mission operations, with applications ranging from multi-robot planetary exploration to agentic AI systems. He holds a B.Sc. and M.Sc. in Photonics and Computer Vision from Darmstadt University of Applied Sciences and a Ph.D. in Computer Engineering from Polytechnique Montréal.
\end{IEEEbiography} 

\begin{IEEEbiography}[{\includegraphics[width=1in,height=1.25in,clip,keepaspectratio]{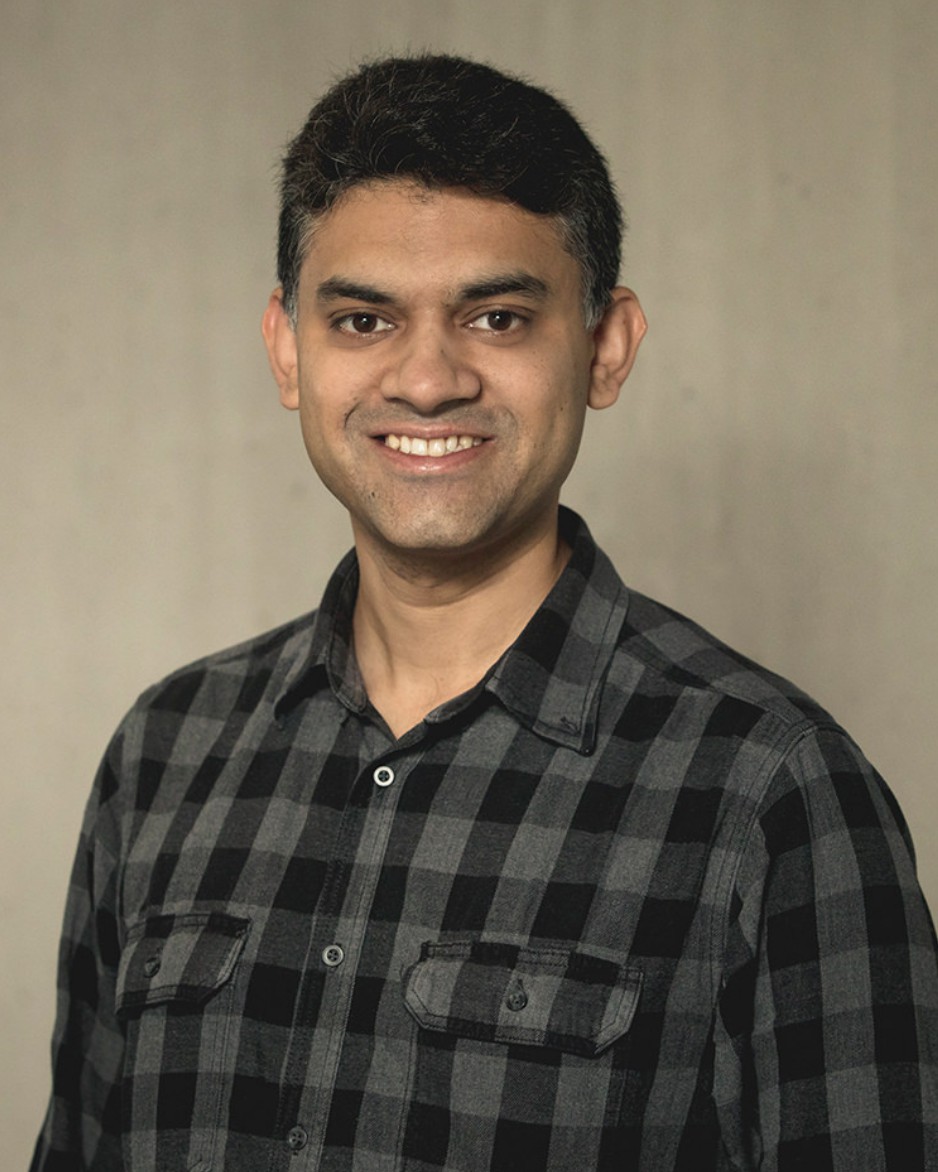}}]
{Shehryar Khattak}~is the Director of Technology at FieldAI, where he leads the FieldAI Wheeled Autonomy Group and serves as the Principal Investigator (PI) for the U.S. Army xTechOverwatch program. Previously, he was a Robotics Technologist in the Perception Systems Group at NASA's Jet Propulsion Laboratory (JPL), where he served as PI for the USSOCOM autonomous multi-robot ISR project and Co-I for the ERNEST lunar nighttime technology demonstrator rover. He also served as perception lead for JPL's team in the DARPA RACER project and was a core contributor to Team CERBERUS, winners of the DARPA Subterranean Challenge. 
\end{IEEEbiography}

\begin{IEEEbiography}[{\includegraphics[width=1in,height=1.25in,clip,keepaspectratio]{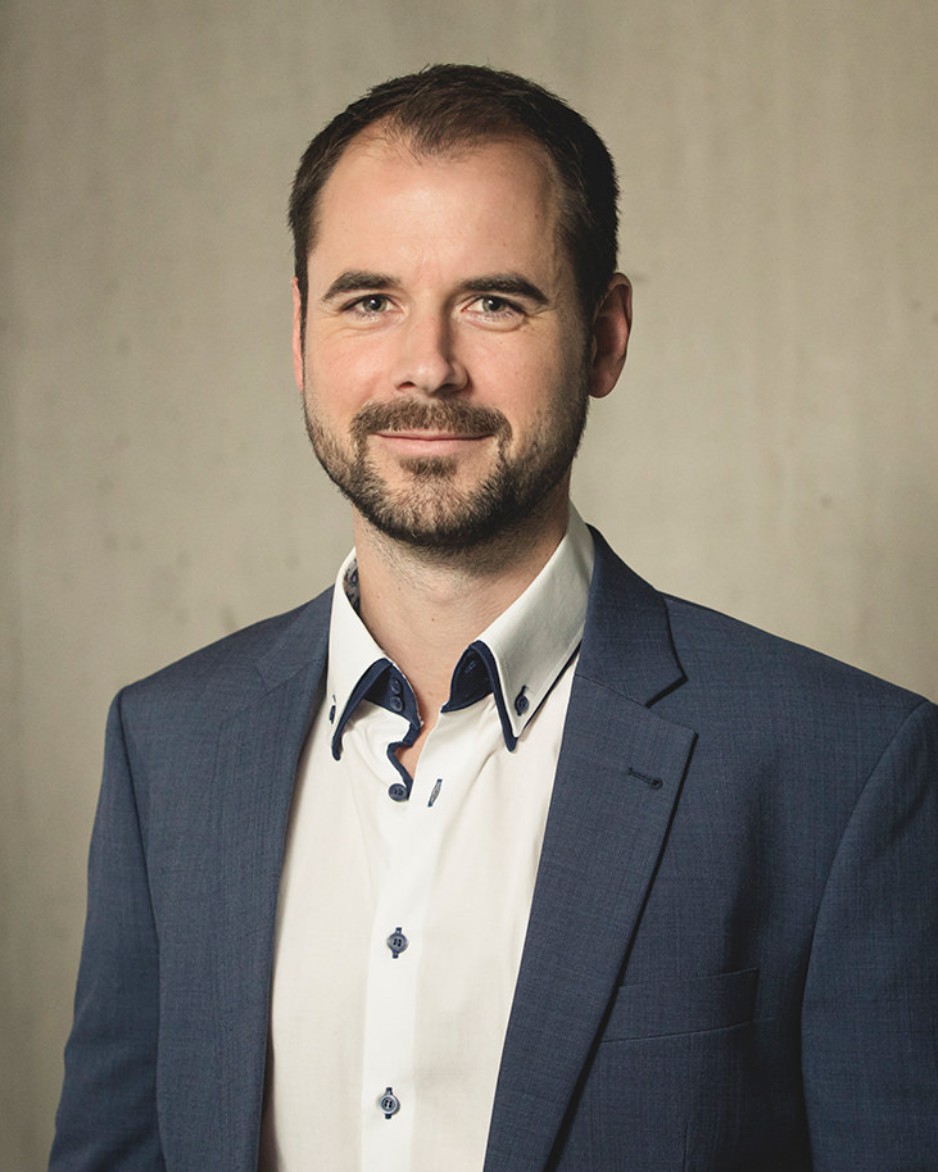}}]{Marco Hutter}~(Member, IEEE) is currently a Professor of robotic systems and the Director of the Center for Robotics, ETH Zürich. He is a Co-Founder of several ETH startups, such as ANYbotics in Zurich, Switzerland, and Gravis Robotics in Zurich. He is also the Director of Boston Dynamics AI Institute Zurich Office, Zurich. He is the Principal Investigator of the NCCRs robotics, automation, and digital fabrication. His research interests include the development of novel machines and actuation concepts, along with the underlying control, planning, and machine learning algorithms for locomotion and manipulation. Dr. Hutter was a recipient of the ERC Starting Grant and the winner of the DARPA SubT Challenge. He received the M.Sc. and Ph.D. degrees in design, actuation, and control of legged robots from the Swiss Federal Institute of Technology (ETH Zürich), Zürich, Switzerland, in 2009 and 2013, respectively.

\end{IEEEbiography} 

\begin{IEEEbiography}[{\includegraphics[width=1in,height=1.25in,clip,keepaspectratio]{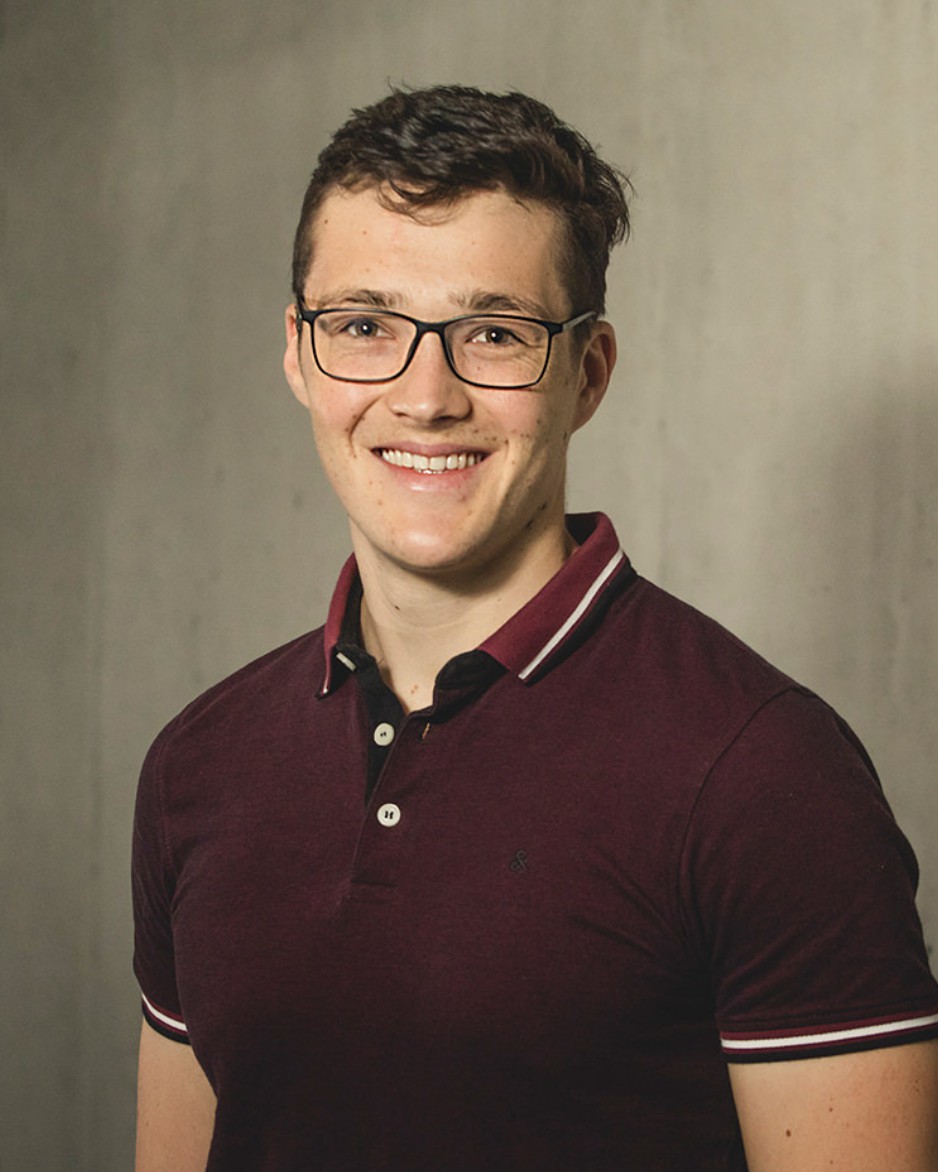}}]
{Jonas Frey}~(Member, IEEE)~is a Postdoctoral Researcher at UC Berkeley and an Affiliate Assistant Professor at the University of Washington. He received his Ph.D. from the Robotic Systems Lab at ETH Zurich and the Max Planck Institute for Intelligent Systems through the MPI ETH Center for Learning Systems. He earned his M.Sc. in Robotics, Systems \& Control in 2021 from ETH Zurich. His research interests focus on perception, navigation, and locomotion to enable the robust deployment of mobile robotic systems.
\end{IEEEbiography} 

\begin{IEEEbiography}[{\includegraphics[width=1in,height=1.25in,clip,keepaspectratio]{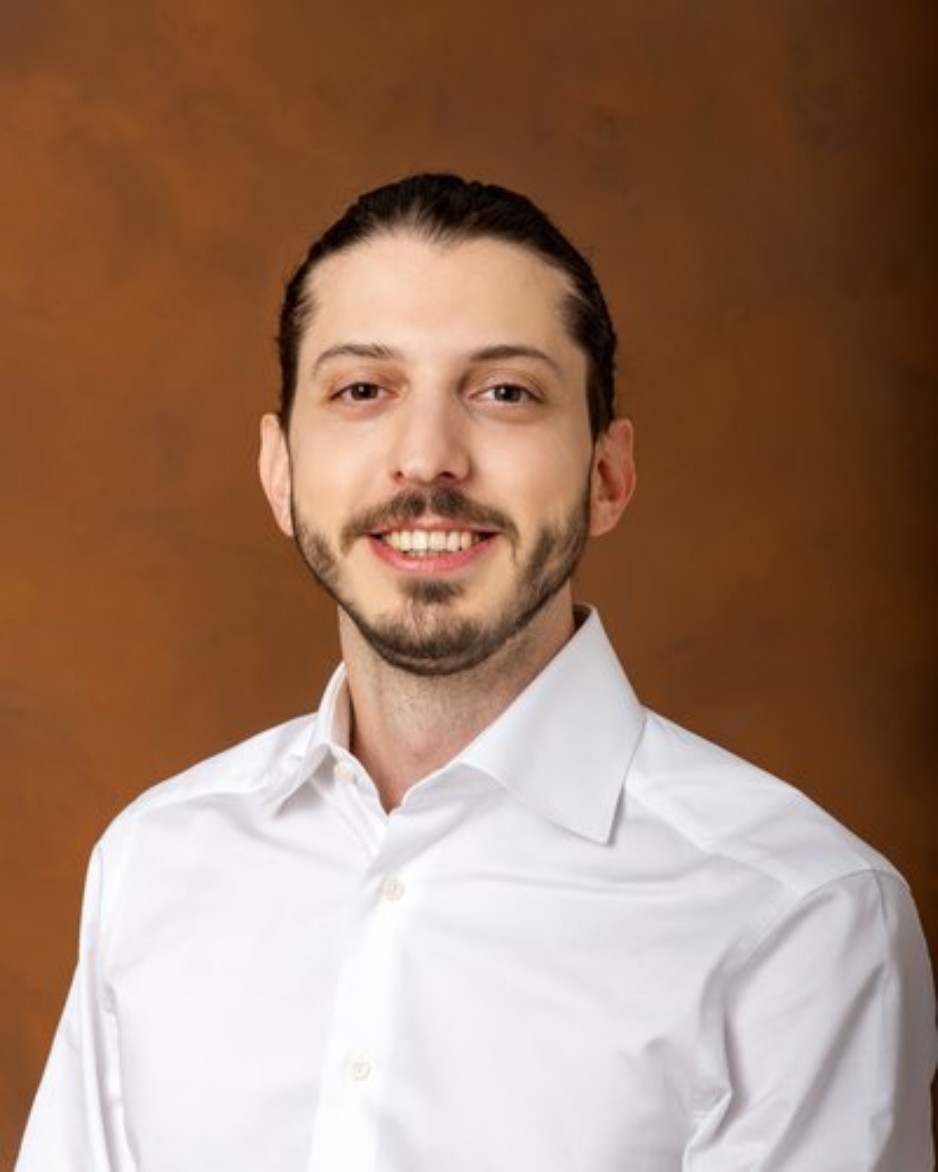}}]
{Patrick Spieler}~(Member, IEEE)~is a Robotics Technologist within the Aerial Robotic Systems Group at NASA Jet Propulsion Laboratory. He received his M.S. (2017) in Robotics and Autonomous Systems and B.S. (2014) in Microengineering from the Swiss Federal Institute of Technology, Lausanne (EPFL). Currently, he is the principal investigator of the NeBula multi-robot autonomy at JPL. Previously he was the PI of JPL's DARPA RACER team. Before JPL, he was a research engineer in the Autonomous Robotics and Control Lab (ARCL) at Caltech where he led the Autonomous Flying Ambulance project and Leonardo, the first flying-walking robot.
\end{IEEEbiography} 

\vfill\pagebreak

\end{document}